\documentclass[final,5p,times,twocolumn,authoryear]{elsarticle}

\usepackage{amssymb}
\usepackage{algorithm}
\usepackage{algpseudocode}
\usepackage{graphicx}

\newcommand{\iee}{{\emph{i.e.},} }

\usepackage{subfigure}
\usepackage{amsmath}
\usepackage{amssymb}
\usepackage{xcolor}
\usepackage{multirow}
\usepackage{tabu}
\usepackage{pbox}
\usepackage{amssymb}
\usepackage{amsfonts}
\usepackage{bbm}
\usepackage{graphicx}
{\begin{list}               %    with flush left bullets.
    {$\bullet$ \hfill}{
        \setlength{\leftmargin}{\parindent}
        \setlength{\parsep}{0.04\baselineskip}
        \setlength{\itemsep}{0.5\parsep}
        \setlength{\labelwidth}{\leftmargin}
        \setlength{\labelsep}{0em}}
    }
{\end{list}}

\providecommand{\eref}[1]{Eq. \eqref{#1}}  % call \eqref from amstex
\providecommand{\cref}[1]{Chapter~\ref{#1}}
\providecommand{\sref}[1]{Section~\ref{#1}}
\providecommand{\fref}[1]{Figure~\ref{#1}}
\providecommand{\tref}[1]{Table~\ref{#1}}

\providecommand{\norm}[1]{\lVert#1\rVert}

\renewcommand{\vec}[1]{\ensuremath{\boldsymbol{#1}}}
\providecommand{\mat}[1]{\ensuremath{\boldsymbol{#1}}}

\providecommand{\calL}{\mathcal{L}}

\providecommand{\mF}{\mat{F}}

\providecommand{\mI}{\mat{I}}

\providecommand{\mP}{\mat{P}}

\providecommand{\mS}{\mat{S}}

\providecommand{\mW}{\mat{W}}

\providecommand{\mY}{\mat{Y}}

\providecommand{\vf}{\vec{f}}

\providecommand{\vp}{\vec{p}}

\providecommand{\vy}{\vec{y}}

\newcommand{\argmax}[1]{\mathop{\underset{#1}{\mbox{argmax}}}}

\journal{Neurocomputing}

\begin{document}

\begin{frontmatter}

\title{Enhancing Long-Term Person Re-Identification Using \\ Global, Local Body Part, and Head Streams}

\author[1]{Duy Tran Thanh}
\ead{duy.tranthanh@seoultech.ac.kr}

\author[2]{Yeejin Lee}
\ead{yeejinlee@seoultech.ac.kr}

\author[1]{Byeongkeun Kang\corref{cor1}}
\ead{byeongkeun.kang@seoultech.ac.kr}

\cortext[cor1]{Corresponding author.}
\affiliation[1]{organization={Department of Electronic Engineering, Seoul National University of Science and Technology},
            addressline={232 Gongneung-ro, Nowon-gu}, 
            city={Seoul},
            postcode={01811}, 
            country={South Korea}}
\affiliation[2]{organization={Department of Electrical and Information Engineering, Seoul National University of Science and Technology},
            addressline={232 Gongneung-ro, Nowon-gu}, 
            city={Seoul},
            postcode={01811}, 
            country={South Korea}}

\begin{abstract}
This work addresses the task of long-term person re-identification. Typically, person re-identification assumes that people do not change their clothes, which limits its applications to short-term scenarios. To overcome this limitation, we investigate long-term person re-identification, which considers both clothes-changing and clothes-consistent scenarios. In this paper, we propose a novel framework that effectively learns and utilizes both global and local information. The proposed framework consists of three streams: global, local body part, and head streams. The global and head streams encode identity-relevant information from an entire image and a cropped image of the head region, respectively. Both streams encode the most distinct, less distinct, and average features using the combinations of adversarial erasing, max pooling, and average pooling. The local body part stream extracts identity-related information for each body part, allowing it to be compared with the same body part from another image. Since body part annotations are not available in re-identification datasets, pseudo-labels are generated using clustering. These labels are then utilized to train a body part segmentation head in the local body part stream. The proposed framework is trained by backpropagating the weighted summation of the identity classification loss, the pair-based loss, and the pseudo body part segmentation loss. To demonstrate the effectiveness of the proposed method, we conducted experiments on three publicly available datasets (Celeb-reID, PRCC, and VC-Clothes). The experimental results demonstrate that the proposed method outperforms the previous state-of-the-art method.
\end{abstract}


\end{frontmatter}

\section{INTRODUCTION}
\label{sec:introduction}
Person re-identification aims to retrieve images of a specific person described by a query image. In contrast to person tracking across multiple cameras, which requires overlapping areas between cameras to track a person~\citep{dukeMTMC2016}, person re-identification is not bound by this constraint. Consequently, it can retrieve images even if they were captured at different locations and times compared to the query image. This characteristic makes person re-identification well-suited for various applications that do not involve overlapping cameras, such as surveillance camera systems and service robots~\citep{person_reID_Neurocomputing_Miao_2023, person_reID_Neurocomputing_Han_2023, person_reID_Neurocomputing_Leng_2023, person_reID_Neurocomputing_Yin_2022}.

\begin{figure*}[!t] \begin{center}
\begin{minipage}{0.98\linewidth}
\centerline{\includegraphics[scale=0.55]{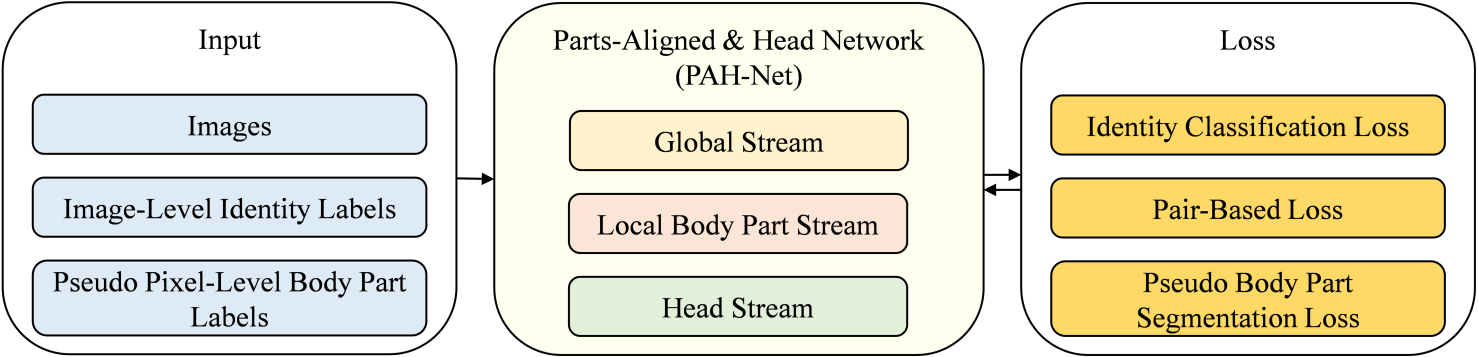}}
\end{minipage}
   \caption{Illustration of the proposed framework. The proposed Parts-Aligned and Head Network (PAH-Net) consists of three streams: the global, local body part, and head streams. The network is trained by backpropagating the combination of three losses: identity classification loss, pair-based loss, and body part segmentation loss. During training, pseudo pixel-level body part labels are generated and utilized along with training images and their image-level identity labels.}
\label{fig:overview}
\end{center}\end{figure*}

In an intelligent surveillance camera system, person re-identification is used to track suspicious, missing, or significant individuals across non-overlapping camera networks. Similarly, a service robot can utilize person re-identification to provide tailored services to specific individuals, regardless of whether they relocate or appear at different times.

Compared with person recognition/identification~\citep{person_identification_EAAI_2023, personRecognition2020}, person re-identification is more practical because it does not require a training dataset containing images of a specific target person. Since a person recognition system aims to identify a person's identity based on an image, it necessitates a training dataset with images of the target person. In contrast, a person re-identification algorithm learns effective methods to determine whether the person in an image matches the person in a query image. Consequently, a trained person re-identification model can be applied to images of any identity, not limited to the persons in the training dataset.

Typical person re-identification methods operate under the assumption that individuals do not change their clothes~\citep{person_reID_Neurocomputing_Song_2021}. Consequently, datasets for this task contain images of each identity solely wearing the same clothes~\citep{cuhk03, zheng2015scalable}. Moreover, since most datasets consist of images cropped by a person detection algorithm, clothing occupies a substantial portion of each image. Consequently, learning-based methods implicitly heavily rely on clothing information. This reliance poses limitations for these methods when the target persons intentionally or naturally change their clothes.

To overcome this limitation, researchers have introduced a more challenging task known as long-term person re-identification, also referred to as clothes-changing person re-identification~\citep{huang2019beyond, huang2019celebrities, yang2019person, wan2020person}. Successfully solving this problem is essential for reliably re-identifying persons across multiple days, even when they intentionally or naturally change their clothes. Datasets designed for this task typically include both clothes-consistent and clothes-changing images, enabling the developed methods to be applicable in both scenarios~\citep{huang2019beyond, yang2019person, wan2020person}.

Considering its importance, we present a novel and effective framework for long-term person re-identification. Firstly, to effectively compare the person in a query image with individuals in gallery images, it is crucial to juxtapose the person's information with the corresponding details of others (\iee comparing the eye color of the person with those of others). To address this, we introduce a local body part stream that aims to encode meaningful information for each body part. During training, we employ a clustering algorithm for implicit body part segmentation. This stream enables the comparison of features from a specific body part in a query image to those of the same body part in gallery images.
Moreover, recognizing the importance of face information in long-term person re-identification due to its lower variance compared with clothing information, we introduce a head stream in addition to the implicit local body part stream. This stream explicitly detects and crops the head region in an image. It then extracts identity-relevant information specifically from the head region.
Lastly, we introduce a global stream that aims to learn and extract identity-relevant information from the entire image.

By integrating the three streams: the global, local body part, and head streams (see~\fref{fig:overview}), we propose a comprehensive framework that effectively learns and leverages both global and local information. We experimentally demonstrate the complementary nature of these three streams as well as the significance of the explicit head and implicit local body part streams in~\sref{sec:result_analysis}.

During training, the proposed framework is trained by backpropagating an identity classification loss, a pair-based loss, and a body part segmentation loss. We demonstrate the effectiveness of the proposed method by conducting experiments on three publicly available datasets: Celeb-reID~\citep{huang2019beyond}, PRCC~\citep{yang2019person}, and VC-Clothes~\citep{wan2020person}.

The contributions of this paper are as follows:
\begin{itemize}
  \item We introduce an effective framework for long-term person re-identification, that learns and extracts both global and local information. The framework consists of three streams: one dedicated to encoding global information and the other two focused on extracting local features. The global stream extracts global information, while the local body part stream and the head stream encode local features.
  \item In the global and head streams, we encode three feature vectors that encompass the most distinct, less distinct, and average features. This encoding is achieved through a combination of adversarial erasing, average pooling, and max pooling techniques.
  \item To encode local information, we utilize both explicit and implicit approaches. The local body part stream utilizes a clustering algorithm to implicitly discover body parts, while the head stream employs an explicit head detection method.
  \item We demonstrate the superior performance of the proposed method compared with the previous state-of-the-art methods through experiments conducted on three publicly available datasets.
\end{itemize}

\section{RELATED WORKS}
\subsection{Long-Term Person Re-Identification}
To accurately retrieve images of the corresponding person in a gallery dataset regardless of varying clothes, researchers have proposed various methods for long-term (clothes-changing) person re-identification. These approaches encompass feature distillation, color information elimination, gait information utilization, feature alignment, and the introduction of novel layers/structures.

Feature distillation is commonly employed to disentangle identity-relevant features from identity-irrelevant features. This technique often involves image reconstruction and adversarial losses, along with swapping feature vectors. \cite{qian2020long} proposed a method consisting of a shape embedding module and two cloth-elimination shape-distillation (CESD) modules. The CESD module aims to disentangle features into cloth-related and cloth-irrelevant components using a self-attention mechanism. To achieve this, it leverages clothes labels along with identity labels in the LTCC dataset~\citep{qian2020long}. However, person re-identification datasets typically lack clothes labels. Therefore, alternative methods have been proposed that do not depend on clothes labels for feature distillation. \cite{li2021learning} introduced the clothing-agnostic shape extraction network (CASE-Net), which comprises a shape encoder and a color encoder. The shape encoder extracts body shape representations that are invariant to clothing changes, while the color encoder encodes clothes-relevant color information. Additionally, similar to~\citep{dgnet2019}, grayscale images are used to guide feature distillation further. \cite{gu2022clothes} proposed a framework consisting of an identity classifier and a clothes classifier. They first train only the clothes classifier using clothes labels and then optimize the identity classifier using the frozen clothes classifier. This compels the backbone to extract clothes-irrelevant features.

As clothing changes can lead to significant variations in color information, researchers have also proposed methods to extract and utilize shape information. This is because the structural aspects of each identity, such as the ratio of body parts, are less affected by attire compared to color information. As mentioned earlier, \cite{qian2020long} proposed a method that includes a shape embedding module. This module explicitly estimates body pose and encodes relationships between body keypoints. \cite{hong2021fine} introduced a two-stream framework called fine-grained shape-appearance mutual learning (FSAM), which consists of an appearance stream and a shape stream. While the shape embedding module in~\citep{qian2020long} utilizes body keypoints, the shape stream in this work employs foreground masks predicted from color images. The appearance stream aims to extract face and hairstyle features, which complement the body shape features from the shape stream. Mutual learning is applied to transfer knowledge from the shape stream to the appearance stream. In the inference phase, only the appearance stream is utilized.

To effectively compare a person in a query image with individuals in gallery images, researchers have proposed methods that involve comparing features of specific body parts between the person in the query image and the individuals in the gallery images. These methods often utilize human pose estimation, human parsing (body part segmentation), and/or attention mechanisms~\citep{huang2019celebrities, wan2020person}. Regarding methods utilizing human pose estimation, \cite{huang2019celebrities} introduced a two-stream neural network-based approach that incorporates a pose estimation algorithm~\citep{openpose2021}. Predicted body keypoints are used to crop the input image into five overlapping segments. These cropped images represent the head region, the area above the waist, the area below the neck, the area below the waist, and the region between the neck and waist. These segments, along with the entire image, are utilized to enhance accuracy. \cite{wan2020person} proposed a three-stream network named the appearance, part, and face extractor network (3APF), which also utilizes a pose estimation algorithm~\citep{openpose2021}. Unlike the method in~\citep{huang2019celebrities}, the 3APF network employs intermediate pose heat maps instead of body keypoints to obtain part-aligned features, resembling the approach in~\citep{suh2018part}. Additionally, unlike the fixed parameters in the previous work, the network parameters in 3APF are dynamically updated using a re-identification loss. Besides part-aligned features, the 3APF network incorporates face information by explicitly detecting the face region using PyramidBox~\citep{pyramidBox2018}.

Other previous works have explored the utilization of advanced layers/structures, incorporation of gait information, and addressing slightly different problems within the context of person re-identification. Concerning the use of networks other than typical convolutional neural networks, \cite{huang2019beyond} introduced the ReIDCaps network, based on the capsule network~\citep{capsuleNet2017}. The ReIDCaps network employs vector-neuron capsules to encode different identities using the vector's length and to represent variations in clothing by the vector's orientation. Additionally, \cite{bansal2022cloth} presented a framework based on the vision transformer (ViT)~\citep{dosovitskiy2020image}. Regarding the use of gait information, \cite{jin2022cloth} proposed a two-stream framework comprising an image stream and a gait recognition stream during training. In the inference phase, only the image stream is utilized for efficiency. To extract gait information from a single image, the gait stream first converts the input image to a silhouette image and then predicts the gait sequence to capture motion cues. Addressing a slightly different problem, \cite{yu2020cocas} aimed to retrieve an image that matches both the identity of a query image and the clothes of another given image. They introduced the biometric-clothes network (BC-Net), consisting of a biometric feature branch and a clothes feature branch. The clothes feature branch takes either a clothes template image or a cropped image using a clothes region detector.

The proposed method aims to explicitly utilize face/head information and implicitly incorporate part-aligned features. Similar to previous work in~\citep{wan2020person}, we explicitly crop the head region from an input image to encode identity-relevant information. However, unlike the previous work, we extract three feature vectors to capture the most distinct, less distinct, and average information. This is achieved by employing combinations of adversarial erasing, max pooling, and average pooling. The same strategy is also applied to the global stream of the proposed method. Regarding part-aligned features, previous methods utilize a human pose estimation algorithm~\citep{openpose2021} trained using body keypoint annotations~\citep{huang2019celebrities, wan2020person}. In contrast to~\citep{huang2019celebrities, wan2020person}, our proposed method does not rely on a pre-trained model or human annotations to align features. Instead, we employ a clustering algorithm to match regions of a query image with those of gallery images. While a similar strategy was proposed in clothes-consistent re-identification~\citep{zhu2020identity}, we further demonstrate the applicability of a clustering-based alignment method in the context of long-term re-identification. This approach helps to avoid dependency on human pose estimation or human parsing.

\subsection{Architectures with Multiple Streams in Other Applications}
Neural networks with multiple streams have been employed in various applications to enhance robustness and accuracy by extracting different aspects of input data. These networks can be categorized into three types: those using multiple input data types (e.g., RGB image and depth map), those using input data and its processed information (e.g., RGB image and optical flow), and those with a single data type but featuring parallel paths.

Networks that take multiple types of input data have been used to leverage complementary information from diverse sources. Researchers have explored the fusion of various data types, such as RGB images and depth maps~\citep{rgbdFusion2023, depth2018Kang, incremental2019Nakajima}, RGB images and point clouds~\citep{rgbPointCloudFusion2024}, and RGB images and near-infrared (NIR) images~\citep{rgbNIRFusion2023} for diverse computer vision tasks.

Researchers have also explored neural networks that take both original data and its processed form to achieve enhanced performance. Relevant studies include the utilization of extracted hand-crafted information such as optical flow from traditional algorithms~\citep{rgbOpticalFlow2014} as well as extracted information by neural networks such as outputs from optical flow networks~\citep{rgbOpticalFlow2023} and human pose estimation networks~\citep{huang2019celebrities, wan2020person}.

Networks with parallel paths for a single data type have gained popularity for their high performance by learning diverse representations~\citep{inceptionNet2015}. They have been applied to tasks such as person re-identification~\citep{zhou2019omni}, semantic segmentation~\citep{HRNetv2_2021}, and human pose estimation~\citep{sun2019deep}.

\begin{figure*}[!t] 
\begin{center}
\begin{minipage}{0.98\linewidth}
\centerline{\includegraphics[scale=0.55]{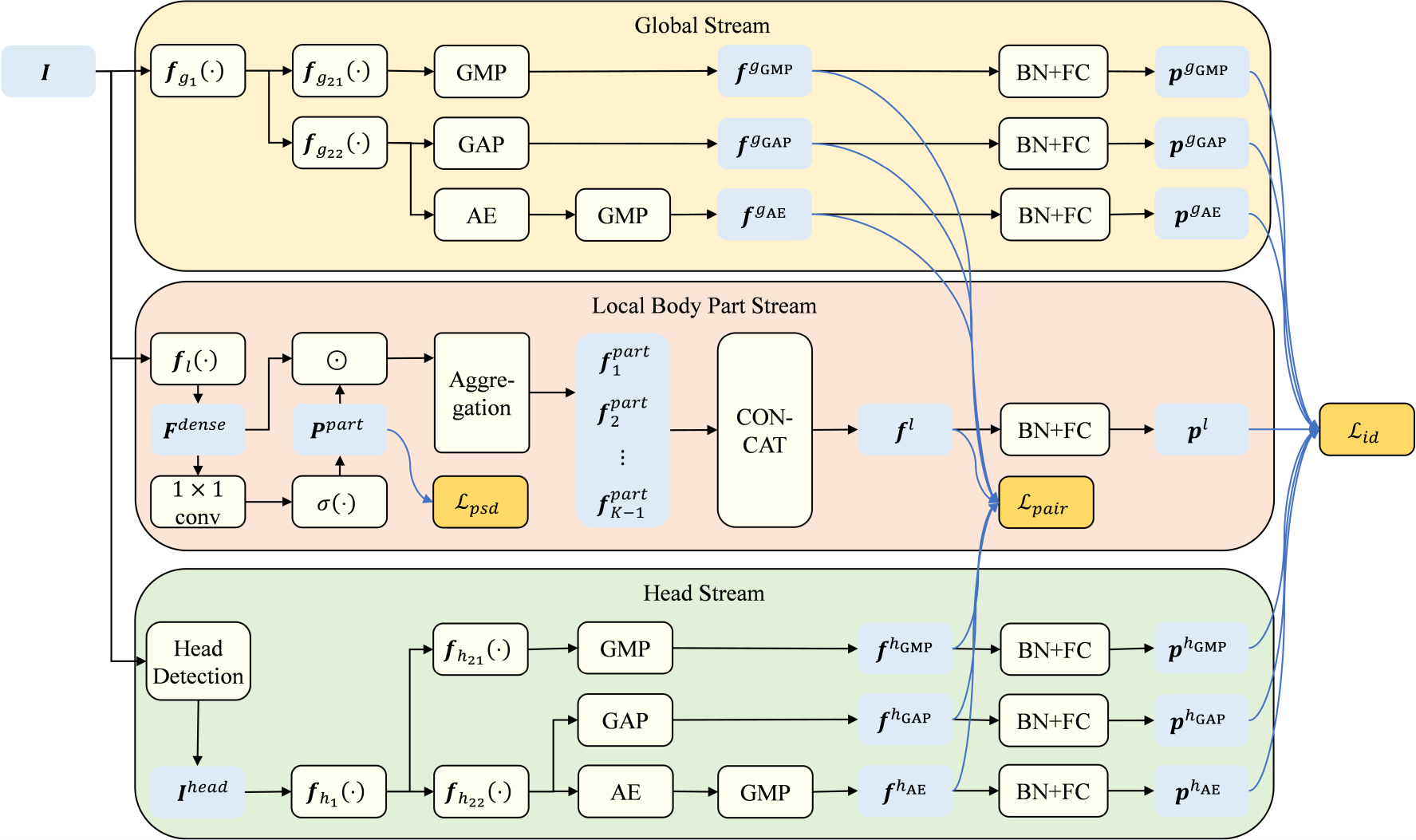}}
\end{minipage}
   \caption{The proposed framework during training. GMP, GAP, and AE denote global max pooling, global average pooling, and adversarial erasing, respectively. BN+FC represents batch normalization followed by a fully connected layer. Aggregation and CONCAT denote feature aggregation over each body part and concatenation, respectively. $\odot$ denotes an element-wise multiplication for each channel. $\calL_{id}$, $\calL_{pair}$, and $\calL_{psd}$ represent the identity classification loss, the pair-based loss, and the pseudo body part segmentation loss, respectively.}
\label{fig:framework_training}
\end{center}
\end{figure*}

The proposed network consists of three streams: global, local body part, and head streams, where all the streams take the same RGB image. While the global stream uses the raw data, the local body part stream aims to estimate implicit body parts intermediately, and the head stream explicitly extracts the head region. Accordingly, the proposed network can be viewed as a neural network that takes input data and its extracted information. It is worth noting that the proposed method extracts this information using parts of the neural network and does not involve any hand-crafted processing steps, as illustrated in~\fref{fig:framework_training}. Moreover, the local body part stream learns to segment implicit body parts by guiding a frontal part of the network using pseudo-labels. Additionally, each stream of the proposed method comprises multiple parallel paths because its backbone is based on either the OSNet backbone~\citep{zhou2019omni} or the HRNet backbone~\citep{sun2019deep}, as shown in Figures~\ref{fig:OSNet_backbone} and~\ref{fig:HRNet_backbone}.

\section{PROPOSED METHOD}
\label{sec:method}
We propose a framework for long-term person re-identification that incorporates three streams: the global stream, the local body part stream, and the head stream, as depicted in~\fref{fig:overview}. The global stream aims to extract identity-relevant features from an entire image. The local body part stream aims to encode features for each body part (local region) to compare the features of a body part from a query image with those of the same body part from gallery images. Because typical re-identification datasets do not provide annotations for body parts, we generate pseudo-labels using a clustering algorithm and utilize them to train a human parsing (body part segmentation) head. The purpose of this parsing network is not to precisely segment each body part but to compare the features of a body part with those of the same body part. The head stream explicitly crops the head region and extracts features from the cropped image. The features from these three streams are combined for re-identification. We refer to the proposed framework as the Parts-Aligned and Head (PAH) network. Detailed information about the proposed framework is presented in~\fref{fig:framework_training}.

\subsection{Network Architecture}
\subsubsection{Global Stream}
\label{sec:global}
The objective of the global stream is to extract identity-relevant features from an image containing a person. Initially, it extracts feature maps using a backbone network based on OSNet~\citep{zhou2019omni}. Subsequently, it encodes the most distinct, less distinct, and average features using a combination of adversarial erasing, max pooling, and average pooling techniques. The parameters of this stream are optimized by backpropagating the identity classification loss and the pair-based loss.

Given an input image $\mI \in \mathbb{R}^{384 \times 128 \times 3}$, this stream first extracts feature representations $\mF^{g_1} \in \mathbb{R}^{48 \times 16 \times 384}$ using the front part of the OSNet architecture~\citep{zhou2019omni}. Specifically, we employ the layers up to the first layer of the third block in OSNet. This process is denoted by $f_{g_1}(\cdot)$ in~\fref{fig:framework_training}. The resulting feature map $\mF^{g_1}$ is then passed through two branches to obtain two distinct feature maps ($\mF^{g_{21}}, \mF^{g_{22}}$). Although the two branches do not share parameters, both utilize the structure of the remaining part of OSNet up to the fifth block. Therefore, both feature maps have the same dimensions of 24$\times$8$\times$512. The operations of the two branches are denoted by $f_{g_{21}}(\cdot)$ and $f_{g_{22}}(\cdot)$, respectively, in~\fref{fig:framework_training}.

\begin{figure}[!t] 
\begin{center}
\begin{minipage}{1\linewidth}
\centerline{\includegraphics[scale=0.48]{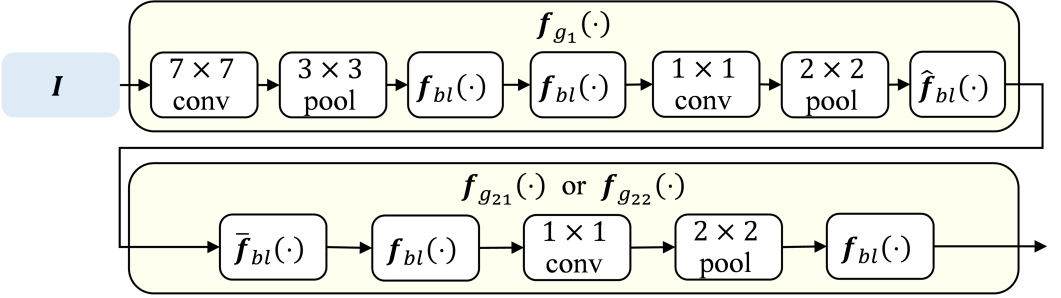}}
\end{minipage}
\begin{minipage}{1\linewidth}
\centerline{(a)}
\end{minipage}
\\
\vspace{0.1cm}
\begin{minipage}{1\linewidth}
\centerline{\includegraphics[scale=0.48]{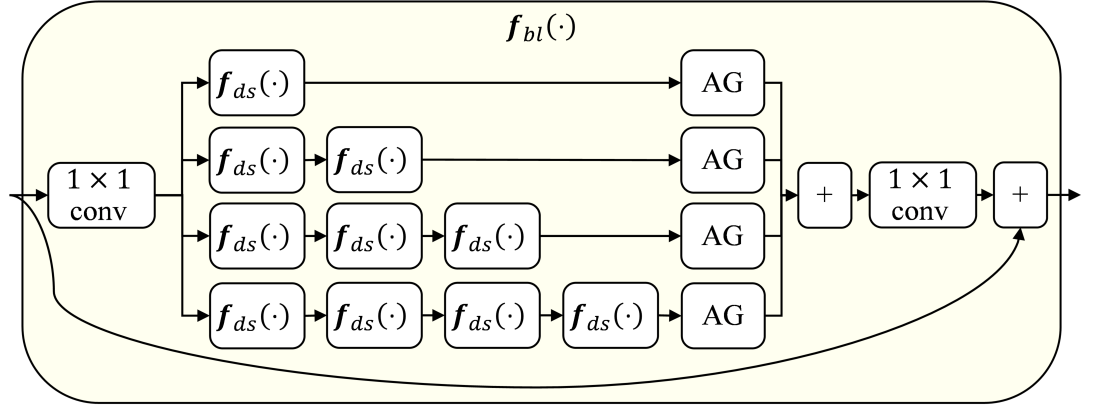}}
\end{minipage}
\begin{minipage}{1\linewidth}
\centerline{(b)}
\end{minipage}
   \caption{Illustration of the networks $f_{g_1}(\cdot)$, $f_{g_{21}}(\cdot)$, and $f_{g_{22}}(\cdot)$ based on the OSNet backbone~\citep{zhou2019omni} given an input image $\mI$. $f_{bl}(\cdot)$ represents a block in OSNet~\citep{zhou2019omni}. $\hat{f}_{bl}(\cdot)$ and $\bar{f}_{bl}(\cdot)$ denote the first layer and the remaining layers of a block, respectively. (a) Structure of the networks $f_{g_1}(\cdot)$, $f_{g_{21}}(\cdot)$, and $f_{g_{22}}(\cdot)$; (b) Structure of a block $f_{bl}(\cdot)$.}
\label{fig:OSNet_backbone}
\end{center}
\end{figure}

More details about the networks $f_{g_1}(\cdot)$, $f_{g_{21}}(\cdot)$, and $f_{g_{22}}(\cdot)$ are illustrated in~\fref{fig:OSNet_backbone} (a). The network $f_{g_1}(\cdot)$ consists of a $7 \times 7$ convolution layer, two blocks, a $1 \times 1$ convolution layer, and the first convolution layer of another block. The networks $f_{g_{21}}(\cdot)$ and $f_{g_{22}}(\cdot)$ have the same structure that comprises the remaining layers of the block, another block, a $1 \times 1$ convolution layer, and an additional block, in sequence. Each block consists of a $1 \times 1$ convolution layer, four parallel branches, aggregation gates denoted as AG, and a $1 \times 1$ convolution layer, as illustrated in~\fref{fig:OSNet_backbone} (b). The parallel branches comprise a varying number of Lite convolution layers $f_{ds}(\cdot)$ that are closely related to depthwise separable convolution layers in~\citep{depthwiseSeparable}. The aggregation gates are used to scale extracted representations before fusing them.

The branch $f_{g_{21}}(\cdot)$ focuses on learning the extraction of the most discriminative feature. To achieve this, the feature map $\mF^{g_{21}}$ undergoes global max pooling, and the resulting pooled feature is utilized for computing the losses. Since global max pooling only considers the highest activation value, the parameters in this branch are optimized to capture the most discriminative feature for re-identification. The output $\vf^{g_{\text{GMP}}} \in \mathbb{R}^{512}$ of this branch is computed as follows:
\begin{equation}
\vf^{g_{\text{GMP}}} = \max_{i,j} \mF^{g_{21}}_{i,j} = \max_{i,j} \Big( f_{g_{21}}\big(f_{g_1}(\mI)\big) \Big)_{i,j}
\end{equation}
where $i$ and $j$ are indexes along height and width dimensions.

The other branch, $f_{g_{22}}(\cdot)$, encodes the less discriminative feature $\vf^{g_{\text{AE}}}$ and the average feature $\vf^{g_{\text{GAP}}}$ vectors using a combination of adversarial erasing, max pooling, and average pooling. The vector $\vf^{g_{\text{GAP}}}$ is obtained by applying global average pooling to the feature map $\mF^{g_{22}}$. Unlike the previous branch $f_{g_{21}}(\cdot)$, the output depends on the values across the entire spatial region. Therefore, it focuses on capturing identity-relevant information from the entire image. The output $\vf^{g_{\text{GAP}}} \in \mathbb{R}^{512}$ is computed as follows:
\begin{equation}
\vf^{g_{\text{GAP}}} = \frac{1}{N} \sum_{i,j} \mF^{g_{22}}_{i,j} = \frac{1}{N} \sum_{i,j} \Big(  f_{g_{22}}\big(f_{g_1}(\mI)\big) \Big)_{i,j}
\end{equation}
where $N$ is the total number of pixels along the spatial dimensions of $\mF^{g_{22}}$.

\begin{figure}[!t] 
\begin{center}
\begin{minipage}{0.22\linewidth}
\centerline{\includegraphics[scale=0.4]{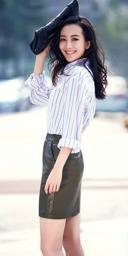}}
\end{minipage}
\begin{minipage}{0.22\linewidth}
\centerline{\includegraphics[scale=0.4]{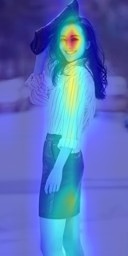}}
\end{minipage}
\begin{minipage}{0.22\linewidth}
\centerline{\includegraphics[scale=0.4]{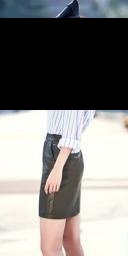}}
\end{minipage}
\\

\begin{minipage}{0.22\linewidth}
\centerline{(a)}
\end{minipage}
\begin{minipage}{0.22\linewidth}
\centerline{(b)}
\end{minipage}
\begin{minipage}{0.22\linewidth}
\centerline{(c)}
\end{minipage}
   \caption{Illustration of adversarial erasing in the global stream. (a) Input image; (b) Sum of squared activation values in Eq. (3) along the channel axis; (c) Adversarially erased input image.}
\label{fig:AE_global}
\end{center}
\end{figure}

While the vector $\vf^{g_{\text{GAP}}}$ depends on the entire region, the values in the vector may still be heavily influenced by the highest activation value, which corresponds to the most discriminative feature. However, relying heavily on the features from the most discriminative region is precarious because the most discriminative feature of an identity in a query image might not appear in some gallery images due to occlusion, differences in viewpoint, and other factors. Accordingly, to reduce dependency on the most discriminative part and increase the focus on other regions, we employ adversarial erasing. Specifically, we explicitly erase the horizontal region with the highest activation value as shown in~\fref{fig:AE_global}, similar to the approach in~\citep{topdropblock2020}. Subsequently, we apply global max pooling. The resulting output, $\vf^{g_{\text{AE}}} \in \mathbb{R}^{512}$, is computed as follows:
\begin{align}
	&\mF^{e_1}_{i,j} = \left\{
	\begin{array}{l l}
	0, \; \text{ if } i = \argmax{i}{\sum_{j,k} \big( \mF^{g_{22}}_{i,j,k}} \big)^2   \\
	1, \; \text{ otherwise } \\ 
	\end{array} \right. \\ 
	&\vf^{g_{\text{AE}}} = \max_{i,j} ( \mF^{g_{22}} \odot \mF^{e_1} )_{i,j}
\end{align}
where $\odot$ denotes element-wise multiplication for each channel. The objective of $\vf^{g_{\text{AE}}}$ is to complement $\vf^{g_{\text{GMP}}}$ and $\vf^{g_{\text{GAP}}}$ by encoding an additional identity-relevant feature vector from a different and less distinct region of an image. \fref{fig:AE_global} illustrates an input image, the corresponding activation map computed using $\mF^{g_{22}}$, and the adversarially erased image. While we erase the feature map $\mF^{g_{22}}$ rather than the input image, we visualize the result of erasing the input image for better visualization. Additionally, we erase one-third of the rows, as illustrated in the figure.

Through the aforementioned processes, we obtain $\vf^{g_{\text{GMP}}}$, $\vf^{g_{\text{GAP}}}$, and $\vf^{g_{\text{AE}}}$. These vectors are first processed by batch normalization to obtain $\tilde{\vf}^{g_{\text{GMP}}}$, $\tilde{\vf}^{g_{\text{GAP}}}$, and $\tilde{\vf}^{g_{\text{AE}}}$, respectively. Then, a fully connected layer is applied to each vector to produce an output vector with dimensions matching the total number of identities in the training dataset. Each value in the vector can be interpreted as the probability of belonging to a specific identity in the training dataset. The outputs of the fully connected layers are denoted by $\vp^{g_{\text{GMP}}}$, $\vp^{g_{\text{GAP}}}$, and $\vp^{g_{\text{AE}}}$, respectively, as illustrated in~\fref{fig:framework_training}.

\subsubsection{Local Body Part Stream} 
\label{sec:local}
The local body part stream aims to compare the features of a specific body part from a query image with those of the same body part from gallery images. It enables a comparison focused on individual body parts while being independent from other body parts or regions. In this stream, we extract dense feature maps and average the features within the region of each body part to obtain a feature vector for that body part. We then concatenate the feature vectors for all body parts and utilize them for re-identification.

Since body part annotations are typically not available in person re-identification datasets, we generate and utilize pseudo-label maps during training, similar to the approach in~\citep{zhu2020identity}. The pseudo-labels are obtained by applying a clustering algorithm to the dense feature maps of each identity.

\begin{figure}[!t] 
\begin{center}
\begin{minipage}{1\linewidth}
\centerline{\includegraphics[scale=0.51]{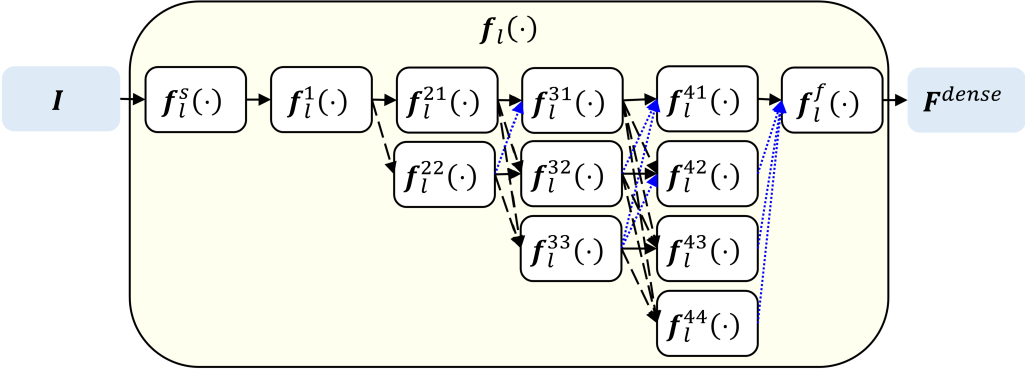}}
\end{minipage}
   \caption{Illustration of the network $f_{l}(\cdot)$ based on the HRNet-W32 backbone~\citep{sun2019deep} to obtain a dense feature representation $\mF^{dense}$ given an input image $\mI$. $f_{l}^{s}(\cdot)$, $f_{l}^{1}(\cdot)$, and $f_{l}^{f}(\cdot)$ represent the stem layers, the first stage, and the final layers, respectively. $f_{l}^{mn}(\cdot)$ denotes the $n$-th parallel branch in the $m$-th stage. $f_{l}^{mn}(\cdot)$ processes the feature maps at a reduced resolution by $1/2^{(n-1)}$ for $n > 1$. The blue dotted line represents bilinear upsampling and a convolution layer. The black solid and dashed lines denote convolution layers with strides of 1 and 2, respectively.}
\label{fig:HRNet_backbone}
\end{center}
\end{figure}

The input image $\mI$ is initially processed using the HRNet-W32 backbone to obtain a dense feature representation $\mF^{dense} \in \mathbb{R}^{96 \times 32 \times 1920}$. We choose the HRNet-W32 backbone over the OSNet backbone~\citep{zhou2019omni} due to its superior performance in tasks that require precise localization, such as human pose estimation, semantic segmentation, and pixel-level attention estimation~\citep{sun2019deep}. The HRNet-W32 backbone comprises two stem layers, four multi-resolution feature extraction and fusion stages, and two final layers. Each stem layer consists of a convolution layer with a stride of 2. Consequently, the two stem layers reduce spatial resolutions from $384 \times 128$ to $96 \times 32$. Regarding the four subsequent stages, the first stage processes at a single resolution while the second, third, and fourth stages process at two, three, and four different resolutions, respectively, using two, three, and four parallel branches, as illustrated in~\fref{fig:HRNet_backbone}. The uppermost branch maintains the resolution of $96 \times 32$ while the lower branches process at reduced resolutions by 1/2, 1/4, and 1/8 using strided 3 × 3 convolutions. The outputs in the parallel branches are fused to exchange multi-resolution information using bilinear upsampling and a convolution layer. After the four stages, we upsample the outputs from the last stage to the resolution ($96 \times 32$) of the uppermost branch and concatenate them. Subsequently, two final convolution layers are applied to the concatenated feature map. For more details about the HRNet-W32 backbone, we refer the readers to the reference~\citep{sun2019deep}. This processing step conducted by the backbone is denoted as $f_{l}(\cdot)$ in~\fref{fig:framework_training}. 

The dense feature map $\mF^{dense}$ is fed into a learned $1 \times 1$ convolution layer to predict body parts or local regions. Subsequently, a softmax function is applied to obtain the probability $\mP^{part}_{i,j}$ for each pixel belonging to each body part. It is computed as follows:
\begin{equation}
\mP^{part}_{i,j} = \sigma \Big({\mW^{part}} \mF^{dense}_{i,j} \Big) 
\end{equation}
where $\mW^{part} \in \mathbb{R}^{K \times 1920}$ represents the matrix containing the parameters of the $1 \times 1$ convolution layer, and $\sigma(\cdot)$ denotes the softmax function. The variable $K$ represents the number of outputs per pixel, corresponding to the number of body parts plus one for the background. Consequently, $\mP^{part} \in \mathbb{R}^{96 \times 32 \times K}$ contains pixel-level probabilities for $K-1$ human body parts as well as the background. The indices $i$ and $j$ denote positions along the height and width dimensions, respectively.

Then, the representation $\vf^{part}_k$ for each body part $k$ is obtained by aggregating the dense feature map $\mF^{dense}$ weighted by the corresponding confidence map $\mP^{part}_{k}$. Specifically, the weighted feature map $\mF^{part}_k \in \mathbb{R}^{96 \times 32 \times 1920}$ for each body part is computed by element-wise multiplication across all channels. 
\begin{equation}
\mF^{part}_k = \mP^{part}_{k} \odot \mF^{dense}
\end{equation}
where $\odot$ denotes element-wise multiplication performed for each channel. $\mF^{part}$ for the background is not computed because the features from the background should not be compared.

Subsequently, the aggregated feature vector $\vf^{part}_k$ is obtained by applying global average pooling to the weighted feature map $\mF^{part}_k$.
\begin{equation}
\vf^{part}_k = \frac{1}{N} \sum_{i,j} \mF^{part}_{i,j,k}
\end{equation}
where $N$ represents the total number of pixels along the spatial dimensions of $\mF^{part}_k$. The vectors $\vf^{part}_k$ are then concatenated and utilized. This concatenated vector is denoted as $\vf^{l}$. Subsequently, $\vf^{l}$ undergoes batch normalization to obtain $\tilde{\vf}^{l}$. Following this, a fully connected layer is applied to $\tilde{\vf}^{l}$ to generate $\vp^{l}$ where the dimension of $\vp^{l}$ matches the total number of identities in the training dataset.

\subsubsection{Head Stream}
The head stream aims to explicitly extract identity-relevant features from the head region of an image. Given an input image containing a person, the head region is explicitly cropped using a human pose estimation algorithm~\citep{guler2018densepose}. The resulting head image is then processed using the same steps as in the global stream described in~\sref{sec:global}, enabling the encoding of the most distinct, less distinct, and average features from the head region.

In more detail, the head region is identified using the DensePose algorithm~\citep{guler2018densepose}. Subsequently, the region is cropped and resized to match the input size of the global stream. The resulting head image $\mI^{head} \in \mathbb{R}^{384 \times 128 \times 3}$ is then fed into the same framework as the global stream. Specifically, the head image is initially processed by the layers up to the first layer of the third block in the OSNet architecture~\citep{zhou2019omni}. The output of the backbone $f_{h_1}(\cdot)$ is then passed through two branches where each branch consists of the layers of the remaining part of OSNet up to the fifth block. Detailed information is illustrated in~\fref{fig:OSNet_backbone}. While the head stream shares the same structure as the global stream, the two streams do not share their learnable parameters. The operations of the two branches are denoted by $f_{h_{21}}(\cdot)$ and $f_{h_{22}}(\cdot)$, respectively, in~\fref{fig:framework_training}.

\begin{figure}[!t] 
\begin{center}
\begin{minipage}{0.22\linewidth}
\centerline{\includegraphics[scale=1.4628]{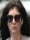}}
\end{minipage}
\begin{minipage}{0.22\linewidth}
\centerline{\includegraphics[scale=1.4628]{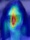}}
\end{minipage}
\begin{minipage}{0.22\linewidth}
\centerline{\includegraphics[scale=1.4628]{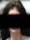}}
\end{minipage}
\\

\begin{minipage}{0.22\linewidth}
\centerline{(a)}
\end{minipage}
\begin{minipage}{0.22\linewidth}
\centerline{(b)}
\end{minipage}
\begin{minipage}{0.22\linewidth}
\centerline{(c)}
\end{minipage}
   \caption{Illustration of adversarial erasing in the head stream. (a) Input image; (b) Sum of squared activation values; (c) Adversarially erased input image.}
\label{fig:AE_head}
\end{center}
\end{figure}

This stream encodes the most distinct $\vf^{h_{\text{GMP}}}$, the less distinct $\vf^{h_{\text{AE}}}$, and the average $\vf^{h_{\text{GAP}}}$ feature vectors from the head image, corresponding to $\vf^{g_{\text{GMP}}}$, $\vf^{g_{\text{AE}}}$, and $\vf^{g_{\text{GAP}}}$ in the global stream, respectively. Each of these feature vectors has a dimension of $512$. Subsequently, each vector is processed by batch normalization and a fully connected layer in sequence, similar to the global stream. The outputs of the batch normalization are represented by $\tilde{\vf}^{h_{\text{GMP}}}$, $\tilde{\vf}^{h_{\text{AE}}}$, and $\tilde{\vf}^{h_{\text{GAP}}}$, respectively. The outputs of the fully connected layers are denoted by $\vp^{h_{\text{GMP}}}$, $\vp^{h_{\text{AE}}}$, and $\vp^{h_{\text{GAP}}}$, respectively, in~\fref{fig:framework_training}. We visualize adversarial erasing in the head stream in~\fref{fig:AE_head}, similar to the global stream in~\fref{fig:AE_global}.

\subsection{Training} 
\label{sec:method_loss}
\subsubsection{Pseudo-Label Generation for Training} 
\label{sec:pseudo_label}
As introduced in~\sref{sec:local}, the local body part stream contains predicting implicit body parts or local regions to obtain a feature vector for each implicit body part. To enable this ability, we generate dense pseudo-label maps for body parts to train the network's parameters which are related to predicting body parts by back-propagating $\calL_{psd}$ in~\fref{fig:framework_training}. During training, the pseudo-label maps are periodically updated.

\begin{figure}[!t] 
\begin{center}
\begin{minipage}{0.22\linewidth}
\centerline{\includegraphics[scale=0.4]{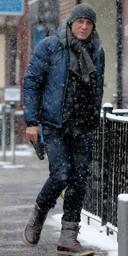}}
\end{minipage}
\begin{minipage}{0.22\linewidth}
\centerline{\includegraphics[scale=0.4]{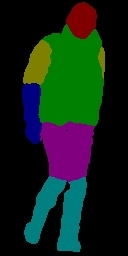}}
\end{minipage}
\begin{minipage}{0.02\linewidth}
\centerline{}
\end{minipage}
\begin{minipage}{0.22\linewidth}
\centerline{\includegraphics[scale=0.4]{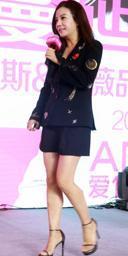}}
\end{minipage}
\begin{minipage}{0.22\linewidth}
\centerline{\includegraphics[scale=0.2]{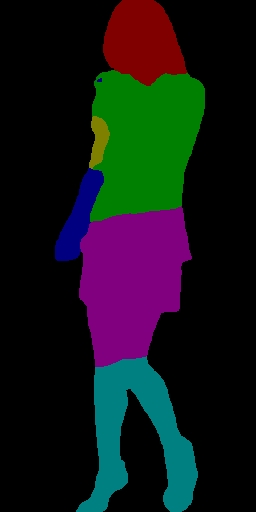}}
\end{minipage}
   \caption{Illustration of generated pseudo-labels using clustering and their corresponding images.}
\label{fig:pseudo_label_generation}
\end{center}
\end{figure}

\fref{fig:pseudo_label_generation} illustrates examples of generated pseudo-label maps along with their corresponding images. These maps are obtained through a two-step clustering process applied to dense feature maps $\mF^{dense}$ extracted from person images. In the first step, pixels are clustered into either foreground or background regions. Subsequently, in the second step, the pixels within the foreground region are further clustered into $K-1$ body parts. 

In the first step, each pixel is assigned to one of two clusters based on the normalized activation values of $\mF^{dense}_{i,j}$ where the two clusters correspond to either the foreground or the background. Specifically, the normalized activation values $\mF^{fg}_{i,j}$ are obtained by dividing the feature map $\mF^{dense}_{i,j}$ by its maximum value across the spatial dimensions. Thus, the normalized activation value $\mF^{fg}_{i,j}$ at pixel ($i,j$) is computed as follows:
\begin{equation}
\mF^{fg}_{i,j} = \frac{\norm{\mF^{dense}_{i,j}}_2}{\max_{i,j} \norm{\mF^{dense}_{i,j}}_2}
\end{equation}
where $\norm{\cdot}_2$ denotes $\ell_2$-norm. $i$ and $j$ are indexes along the height and width dimensions. Subsequently, a clustering algorithm is applied to group pixels ($i,j$) into two clusters based on the normalized activation values $\mF^{fg}_{i,j}$. The cluster with the higher average activation value is then selected as the foreground cluster since the normalized activation values can be approximately interpreted as the probability of being part of the foreground.

In the second stage, pixels in the foreground cluster are categorized into $K-1$ clusters based on the normalized feature vectors extracted from the dense feature map $\mF^{dense}$, where each cluster corresponds to an implicit body part. Specifically, the normalized feature vectors $\mF^{psd}_{i,j}$ are obtained by normalizing each vector $\mF^{dense}_{i,j}$ at pixel ($i,j$) to a unit vector, as shown below:
\begin{equation}
\mF^{psd}_{i,j} = \frac{\mF^{dense}_{i,j}}{\norm{\mF^{dense}_{i,j}}_2}
\end{equation}
where $\norm{\cdot}_2$ represents $\ell_2$-norm. $i$ and $j$ are indexes along the height and width dimensions. Similar to the first step, a clustering algorithm is applied to group pixels ($i,j$) in the foreground cluster into $K-1$ clusters based on $\mF^{psd}_{i,j}$. Due to the unknown associations between clusters and body parts, we assign a pseudo-label to each cluster based on the average vertical position of the pixels within the cluster. It is important to note that we utilize the generated pseudo-labels to train a network rather than using them directly to localize body parts. Consequently, even with occasional noisy labels, the network learns from majority-consistent pseudo-labels and predicts implicit body parts for unseen images. We experimentally demonstrate the effectiveness of the local body part stream in~\sref{sec:results}.

In all experiments, we set $K$ to seven, corresponding to six implicit body parts and one background, as illustrated in~\fref{fig:pseudo_label_generation}. Both clusterings are performed using the $k$-means algorithm. Pseudo-labels are updated at every epoch using the model of that moment, starting from the very first epoch. We use the generated pseudo-label maps to compute a loss in~\eref{eq:loss_part} for training the network. Consequently, clusterings are not applied during inference. The generated pseudo-label maps are denoted as $\mY^{psd} \in \mathbb{Z}^{96 \times 32 \times K}$.

\subsubsection{Loss Function} 
The proposed framework is trained using a loss function $\calL$ that comprises three terms: an identity classification loss $\calL_{id}$, a pair-based loss $\calL_{pair}$, and a pseudo body part segmentation loss $\calL_{psd}$. The identity classification loss is based on the cross-entropy loss, which is commonly used in classification tasks. In person re-identification datasets, images and image-level identity labels are provided for training. Therefore, an image-level identity classification task is formulated to train the network to encode identity-relevant information. While the identity classification loss is computed for each pair of an image and its label, person re-identification involves comparing similarities between gallery images and a query image. Hence, the pair-based loss is additionally employed to consider relative similarities. The pair-based loss maximizes similarities between images of the same identity while minimizing those between images of different identities. It is because person re-identification retrieves the closest image among gallery images given a query image. The pseudo body part segmentation loss is used to train implicit body part segmentation using the generated pseudo-label maps.

In more detail, the identity classification loss $\calL_{id}$ utilizes images and image-level identity labels to train the network to encode identity-relevant information in each feature vector. Accordingly, the loss is computed using the outputs of all the streams, as illustrated in~\fref{fig:framework_training}. The pair-based loss $\calL_{pair}$ uses an anchor image, positive samples containing the same identity as the anchor image, and negative samples consisting of different identities from the anchor image. This loss aims to maximize the similarities between the extracted features of the anchor and positive samples while minimizing the similarities between the vectors of the anchor and negative samples. It is crucial because the most similar images from a gallery set are retrieved based on the similarity metric given a query image during inference. Similar to the identity classification loss, this loss is computed using the outputs of all the streams, as illustrated in~\fref{fig:framework_training}. The pseudo body part segmentation loss $\calL_{psd}$ utilizes the predicted body part probability $\mP^{part}$ in the local body part stream and generated pseudo-labels in~\sref{sec:pseudo_label}. This loss is designed to train the network for implicit body part prediction using pseudo-labels. Rather than using pseudo-labels directly for feature aggregation for each body part, we train the network to learn from majority-consistent pseudo-labels while disregarding occasional noisy labels. Unlike the other two losses, this loss guides only the parameters in the local body part stream, as illustrated in~\fref{fig:framework_training}.

\noindent \textbf{Identity classification loss $\calL_{id}$}. It is computed using the image-level identity label and the network outputs for a given image. Since this loss is designed to guide the network to encode identity-relevant information, the outputs from all streams are utilized to compute this loss. Specifically, it involves the outputs from the last fully connected layers of the global, local body part, and head streams. The set of outputs for the image $\mI$ is represented as follows:
\begin{equation}
\mS^{id} = \Big\{ \vp^{g_{\text{GMP}}}, \vp^{g_{\text{AE}}}, \vp^{g_{\text{GAP}}}, \vp^{l}, \vp^{h_{\text{GMP}}}, \vp^{h_{\text{AE}}}, \vp^{h_{\text{GAP}}} \Big\}.
\end{equation}
Then, the identity classification loss $\calL_{id}$ is computed as follows:
\begin{equation}
\calL_{id}(\mI, \vy) = \sum_{\vp \in \mS^{id}} \calL_{ce}(\vp, \vy)  
				= - \sum_{\vp \in \mS^{id}} \sum_{i=1}^C \vy_i \ln \vp_i
\label{eq:loss_id}
\end{equation}
where $\calL_{ce}(\cdot)$ and $\vy$ represent the cross-entropy loss function and the image-level identity label, respectively. $\vy_i$ takes the value of one if $\mI$ contains the person corresponding to label $i$, and zero otherwise. $C$ denotes the total number of identities in the training dataset.

\noindent \textbf{Pair-based loss $\calL_{pair}$}. It is derived from the multi-similarity loss~\citep{wang_2019_CVPR}. It is computed using a pair of an anchor $\mI$, positive samples $\mS^{pos}=\{\mI^{pos}\}$, and negative samples $\mS^{neg}=\{\mI^{neg}\}$. Positive samples refer to images with the same identity, while negative samples refer to images with different identities. This loss is to maximize the similarities between the feature vectors of the anchor and positive samples, while minimizing the similarities between the vectors of the anchor and negative samples. Similar to the identity classification loss, this loss is applied to all the streams, as the outputs of all the streams are utilized together during inference. Specifically, the loss is computed using the set of vectors before the batch normalization in all the branches. The set of vectors for the image $\mI$ is as follows:
\begin{equation}
\mS^{pair} = \Big\{ \vf^{g_{\text{GMP}}}, \vf^{g_{\text{AE}}}, \vf^{g_{\text{GAP}}}, \vf^{l}, \vf^{h_{\text{GMP}}}, \vf^{h_{\text{AE}}}, \vf^{h_{\text{GAP}}} \Big\}.
\end{equation}
The last fully connected layer is responsible for adjusting the number of outputs to match the number of identities in the training dataset, enabling the prediction of the probability of each identity. As a result, $\calL_{id}$ utilizes the output of the last fully connected layer. On the other hand, $\calL_{pair}$ uses the outputs before this layer, which have dimensions that are independent of the training dataset. The pair-based loss $\calL_{pair}$ is computed as follows:
\begin{equation}
\begin{split}
&\calL_{pair}(\mI, \mS^{pos}, \mS^{neg})= \sum_{\vf \in \mS^{pair}} \calL_{ms}(\vf, \bar{\mS}^{pos}, \bar{\mS}^{neg}) \\
& = \sum_{\vf \in \mS^{pair}} \biggl[ \frac{1}{\alpha_1} \ln \Big\{ 1+\sum_{\vf^{pos} \in \bar{\mS}^{pos}} e^{- \alpha_1 (d_{\vf}(\vf, \vf^{pos}) - \lambda) } \Big\} \\
&\hspace{0.5cm} + \frac{1}{\alpha_2} \ln \Big\{ 1+\sum_{\vf^{neg} \in \bar{\mS}^{neg}} e^{ \alpha_2 ( d_{\vf}(\vf, \vf^{neg})-\lambda) } \Big\} \biggr]
\label{eq:loss_pair}
\end{split}
\end{equation}
where $\bar{\mS}^{pos}$ and $\bar{\mS}^{neg}$ contain the extracted vectors from images in $\mS^{pos}$ and $\mS^{neg}$, respectively, similar to $\mS^{pair}$. $\calL_{ms}(\cdot)$ denotes the multi-similarity loss function, and $d_{\vf}(\vf, \vf^{pos})$ represents the similarity function that calculates the similarity between $\vf$ and the corresponding vector $\vf^{pos}$. The similarity is computed by the scalar product of the two vectors. The margin is denoted by $\lambda$. $\alpha_1$ and $\alpha_2$ are balancing factors used to control the influence of positive and negative samples.

\noindent \textbf{Pseudo body part segmentation loss $\calL_{psd}$}. It is used to train the body part prediction head using pseudo-labels. This loss is computed by summing the cross-entropy losses over the spatial dimensions. Given the predicted probability $\mP^{part}$ for each body part and the generated pseudo-labels $\mY^{psd}$, the loss is computed as follows:
\begin{equation}
\begin{split}
\calL_{psd}(\mI, \mY^{psd}) &= \sum_{i,j} \calL_{ce}(\mP^{part}_{i,j}, \mY^{psd}_{i,j}) \\
&= - \sum_{i,j} \sum_{k=1}^K \mY^{psd}_{i,j,k} \ln (\mP^{part}_{i,j,k})
\label{eq:loss_part}
\end{split}
\end{equation}
where $k$ is an index for clusters (body parts and background). $i$ and $j$ are indexes along the height and width dimensions. 

\noindent \textbf{Total loss $\calL$}. It is the weighted sum of the three loss terms: $\calL_{id}(\mI, \vy)$, $\calL_{pair}(\mI, \mS^{pos}, \mS^{neg})$, and $\calL_{psd}(\mI, \mY^{psd})$. It is computed as follows:
\begin{equation}
\begin{split}
&\calL(\mI, \vy, \mS^{pos}, \mS^{neg}, \mY^{psd}) = \calL_{id}(\mI, \vy) \\
& \hspace{0.5cm} + \lambda_{pair} \calL_{pair}(\mI, \mS^{pos}, \mS^{neg}) + \lambda_{psd} \calL_{psd}(\mI, \mY^{psd})
\label{eq:loss_total}
\end{split}
\end{equation}
where $\lambda_{pair}$ and $\lambda_{psd}$ are to balance between loss terms. We set $\lambda_{pair}$ and $\lambda_{psd}$ as one and $0.1$, respectively, in all experiments. 

\subsection{Inference} 
In the inference stage, the vectors extracted from all the streams are concatenated and utilized. Specifically, the vectors before the last fully connected layer are employed, as the last fully connected layer is designed to match the number of outputs with the number of identities in the training dataset. The concatenated vector $\vf^{final}$ is as follows:
\begin{equation}
\vf^{final} = \bigl[ \tilde{\vf}^{g_{\text{GMP}}}, \tilde{\vf}^{g_{\text{AE}}}, \tilde{\vf}^{g_{\text{GAP}}}, \tilde{\vf}^{l}, \tilde{\vf}^{h_{\text{GMP}}}, \tilde{\vf}^{h_{\text{AE}}}, \tilde{\vf}^{h_{\text{GAP}}} \bigr].
\end{equation}
Then, the cosine distance $d$ is computed between the concatenated vector $\vf^{q}$ from a query image and the vectors $\vf^{g}$ from gallery images. 
\begin{equation}
d = \frac{\vf^{q} \cdot \vf^{g}}{\norm{\vf^{q}}_2 \norm{\vf^{g}}_2}
\label{eq:cosine_distance}
\end{equation}
where $\cdot$ denotes the scalar product.

\begin{table*}[!t]
\centering
\begin{minipage}{1\linewidth}
\caption{Quantitative comparison on the Celeb-reID dataset~\citep{huang2019beyond} (\%).}
\label{tab:result_celeb}
\centering
\begin{tabular}{ >{\centering}m{0.39\textwidth}| >{\centering}m{0.13\textwidth}| >{\centering}m{0.13\textwidth}| >{\centering\arraybackslash}m{0.13\textwidth} } 
\hline
Method & Rank-1 & Rank-5 & mAP \\
\hline\hline
PCB~\citep{sun2018beyond} 			& 37.1 & 57.0 & 8.2 \\  %% short-term + ReIDCaps
MLFN~\citep{chang2018multi} 		& 41.4 & 54.7 & 6.0 \\  %% short-term + ReIDCaps
HACNN~\citep{li2018harmonious} 		& 47.6 & 63.3 & 9.5 \\  %% short-term + ReIDCaps
MGN~\citep{wang2018learning} 		& 49.0 & 64.9 & 10.8 \\  %% short-term + ReIDCaps
ReIDCaps~\citep{huang2019beyond} 	& 51.2 & 65.4 & 9.8 \\  %% original paper
AFD-Net~\citep{xu2021adversarial} 	& 52.1 & 66.1 & 10.6 \\  %% original paper
IS-GAN$_{KL}$~\citep{eom2021disentangled} 	& 54.5 & - & 12.8 \\  %% original paper
IS-GAN$_{DC}$~\citep{eom2021disentangled} 	& 54.5 & - & 12.5\\  %% original paper
IS-GAN$_{KL}$+RR~\citep{eom2021disentangled} 	& 54.9 & - & 14.5 \\  %% original paper
IS-GAN$_{DC}$+RR~\citep{eom2021disentangled} 	& 54.0 & - & 14.9 \\  %% original paper
SirNet~\citep{sirNet2022} 					& 56.0 & 70.3 & 14.2 \\  %% original paper
LightMBN~\citep{herzog2021lightweight} 		& 59.2 & 74.5 & 15.2 \\  %% short-term + IRANet
ReIDCaps+~\citep{huang2019beyond} 			& 63.0 & 76.3 & 15.8 \\  %% original paper
IRANet~\citep{shi2022iranet} 		& 64.1 & \underline{78.7} & \underline{19.0} \\  %% original paper
CASE-Net~\citep{li2021learning} 		& \underline{66.4} & 78.1 & 18.2 \\  %% original paper
\hline
\textbf{Proposed} 		& \textbf{69.0} & \textbf{82.7}  & \textbf{23.6}\\
\hline
\end{tabular}
\end{minipage}
\end{table*}

\begin{figure}[!t] \begin{center}
\begin{minipage}{0.140\linewidth}
\centerline{\footnotesize{Method}}
\end{minipage}
\begin{minipage}{0.140\linewidth}
\centerline{\footnotesize{Query}}
\end{minipage}
\begin{minipage}{0.125\linewidth}
\centerline{\footnotesize{Top-1}}
\end{minipage}
\begin{minipage}{0.125\linewidth}
\centerline{\footnotesize{Top-2}}
\end{minipage}
\begin{minipage}{0.125\linewidth}
\centerline{\footnotesize{Top-3}}
\end{minipage}
\begin{minipage}{0.125\linewidth}
\centerline{\footnotesize{Top-4}}
\end{minipage}
\begin{minipage}{0.125\linewidth}
\centerline{\footnotesize{Top-5}}
\end{minipage}
\\
\vspace{0.05cm}
\rule{\linewidth}{0.4pt}
\\
\vspace{0.05cm}
\begin{minipage}{0.140\linewidth}
\centerline{\footnotesize{ReIDCaps+}}
\end{minipage}
\begin{minipage}{0.140\linewidth}
\centerline{\includegraphics[scale=0.23]{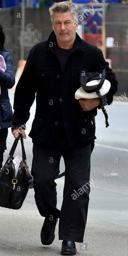}}
\end{minipage}
\begin{minipage}{0.125\linewidth}
\centerline{\includegraphics[scale=0.23]{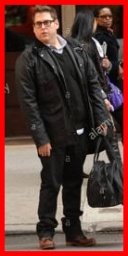}}
\end{minipage}
\begin{minipage}{0.125\linewidth}
\centerline{\includegraphics[scale=0.23]{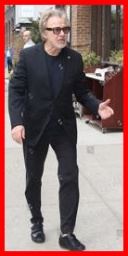}}
\end{minipage}
\begin{minipage}{0.125\linewidth}
\centerline{\includegraphics[scale=0.23]{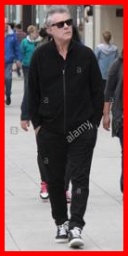}}
\end{minipage}
\begin{minipage}{0.125\linewidth}
\centerline{\includegraphics[scale=0.23]{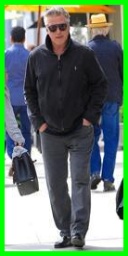}}
\end{minipage}
\begin{minipage}{0.125\linewidth}
\centerline{\includegraphics[scale=0.23]{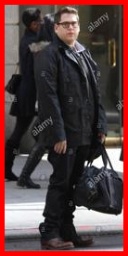}}
\end{minipage}
\\
\vspace{0.05cm}
\begin{minipage}{0.140\linewidth}
\centerline{\footnotesize{SirNet}}
\end{minipage}
\begin{minipage}{0.140\linewidth}
\centerline{}
\end{minipage}
\begin{minipage}{0.125\linewidth}
\centerline{\includegraphics[scale=0.23]{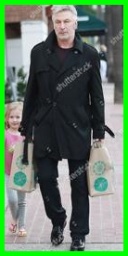}}
\end{minipage}
\begin{minipage}{0.125\linewidth}
\centerline{\includegraphics[scale=0.23]{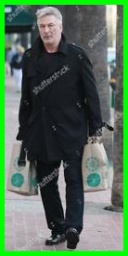}}
\end{minipage}
\begin{minipage}{0.125\linewidth}
\centerline{\includegraphics[scale=0.23]{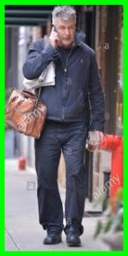}}
\end{minipage}
\begin{minipage}{0.125\linewidth}
\centerline{\includegraphics[scale=0.23]{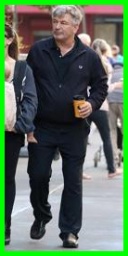}}
\end{minipage}
\begin{minipage}{0.125\linewidth}
\centerline{\includegraphics[scale=0.23]{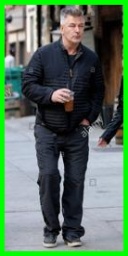}}
\end{minipage}
\\
\vspace{0.05cm}
\begin{minipage}{0.140\linewidth}
\centerline{\footnotesize{Proposed}}
\end{minipage}
\begin{minipage}{0.140\linewidth}
\centerline{}
\end{minipage}
\begin{minipage}{0.125\linewidth}
\centerline{\includegraphics[scale=0.23]{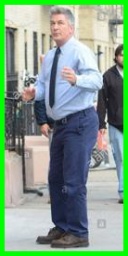}}
\end{minipage}
\begin{minipage}{0.125\linewidth}
\centerline{\includegraphics[scale=0.23]{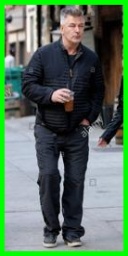}}
\end{minipage}
\begin{minipage}{0.125\linewidth}
\centerline{\includegraphics[scale=0.23]{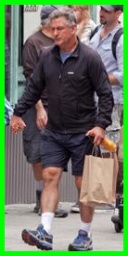}}
\end{minipage}
\begin{minipage}{0.125\linewidth}
\centerline{\includegraphics[scale=0.23]{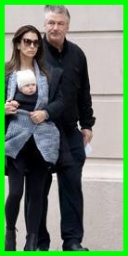}}
\end{minipage}
\begin{minipage}{0.125\linewidth}
\centerline{\includegraphics[scale=0.23]{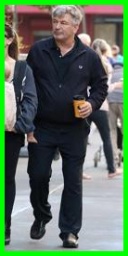}}
\end{minipage}
\\
\vspace{0.05cm}
\rule{\linewidth}{0.4pt}
\\
\vspace{0.05cm}
\begin{minipage}{0.140\linewidth}
\centerline{\footnotesize{ReIDCaps+}}
\end{minipage}
\begin{minipage}{0.140\linewidth}
\centerline{\includegraphics[scale=0.23]{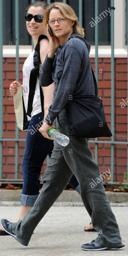}}
\end{minipage}
\begin{minipage}{0.125\linewidth}
\centerline{\includegraphics[scale=0.23]{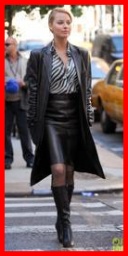}}
\end{minipage}
\begin{minipage}{0.125\linewidth}
\centerline{\includegraphics[scale=0.23]{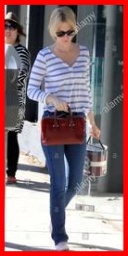}}
\end{minipage}
\begin{minipage}{0.125\linewidth}
\centerline{\includegraphics[scale=0.23]{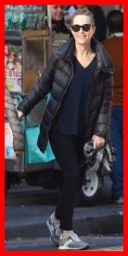}}
\end{minipage}
\begin{minipage}{0.125\linewidth}
\centerline{\includegraphics[scale=0.23]{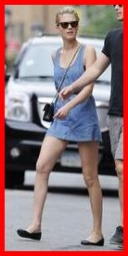}}
\end{minipage}
\begin{minipage}{0.125\linewidth}
\centerline{\includegraphics[scale=0.23]{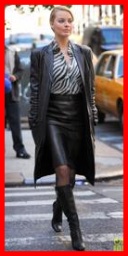}}
\end{minipage}
\\
\vspace{0.05cm}
\begin{minipage}{0.140\linewidth}
\centerline{\footnotesize{SirNet}}
\end{minipage}
\begin{minipage}{0.140\linewidth}
\centerline{}
\end{minipage}
\begin{minipage}{0.125\linewidth}
\centerline{\includegraphics[scale=0.23]{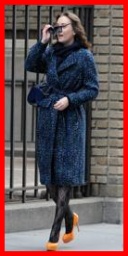}}
\end{minipage}
\begin{minipage}{0.125\linewidth}
\centerline{\includegraphics[scale=0.23]{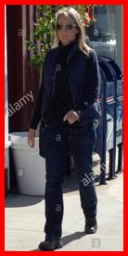}}
\end{minipage}
\begin{minipage}{0.125\linewidth}
\centerline{\includegraphics[scale=0.23]{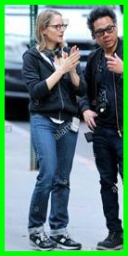}}
\end{minipage}
\begin{minipage}{0.125\linewidth}
\centerline{\includegraphics[scale=0.23]{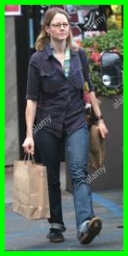}}
\end{minipage}
\begin{minipage}{0.125\linewidth}
\centerline{\includegraphics[scale=0.23]{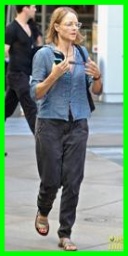}}
\end{minipage}
\\
\vspace{0.05cm}
\begin{minipage}{0.140\linewidth}
\centerline{\footnotesize{Proposed}}
\end{minipage}
\begin{minipage}{0.140\linewidth}
\centerline{}
\end{minipage}
\begin{minipage}{0.125\linewidth}
\centerline{\includegraphics[scale=0.23]{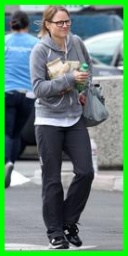}}
\end{minipage}
\begin{minipage}{0.125\linewidth}
\centerline{\includegraphics[scale=0.23]{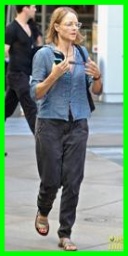}}
\end{minipage}
\begin{minipage}{0.125\linewidth}
\centerline{\includegraphics[scale=0.23]{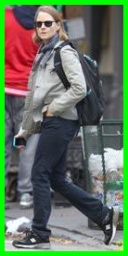}}
\end{minipage}
\begin{minipage}{0.125\linewidth}
\centerline{\includegraphics[scale=0.23]{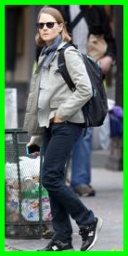}}
\end{minipage}
\begin{minipage}{0.125\linewidth}
\centerline{\includegraphics[scale=0.23]{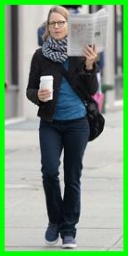}}
\end{minipage}
\\
\vspace{0.05cm}
\rule{\linewidth}{0.4pt}
\\
\vspace{0.05cm}
\begin{minipage}{0.140\linewidth}
\centerline{\footnotesize{ReIDCaps+}}
\end{minipage}
\begin{minipage}{0.140\linewidth}
\centerline{\includegraphics[scale=0.23]{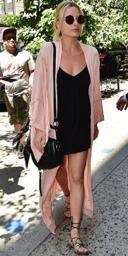}}
\end{minipage}
\begin{minipage}{0.125\linewidth}
\centerline{\includegraphics[scale=0.23]{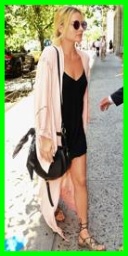}}
\end{minipage}
\begin{minipage}{0.125\linewidth}
\centerline{\includegraphics[scale=0.23]{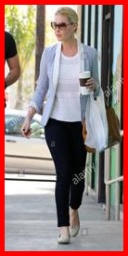}}
\end{minipage}
\begin{minipage}{0.125\linewidth}
\centerline{\includegraphics[scale=0.23]{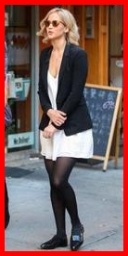}}
\end{minipage}
\begin{minipage}{0.125\linewidth}
\centerline{\includegraphics[scale=0.23]{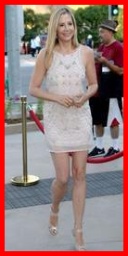}}
\end{minipage}
\begin{minipage}{0.125\linewidth}
\centerline{\includegraphics[scale=0.23]{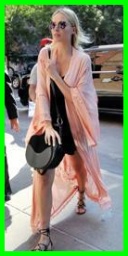}}
\end{minipage}
\\
\vspace{0.05cm}
\begin{minipage}{0.140\linewidth}
\centerline{\footnotesize{SirNet}}
\end{minipage}
\begin{minipage}{0.140\linewidth}
\centerline{}
\end{minipage}
\begin{minipage}{0.125\linewidth}
\centerline{\includegraphics[scale=0.23]{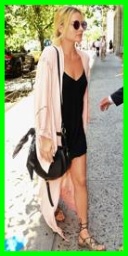}}
\end{minipage}
\begin{minipage}{0.125\linewidth}
\centerline{\includegraphics[scale=0.23]{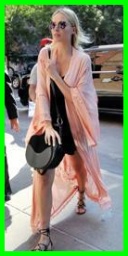}}
\end{minipage}
\begin{minipage}{0.125\linewidth}
\centerline{\includegraphics[scale=0.23]{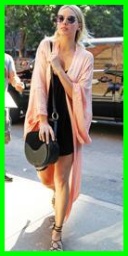}}
\end{minipage}
\begin{minipage}{0.125\linewidth}
\centerline{\includegraphics[scale=0.23]{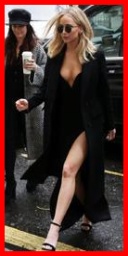}}
\end{minipage}
\begin{minipage}{0.125\linewidth}
\centerline{\includegraphics[scale=0.23]{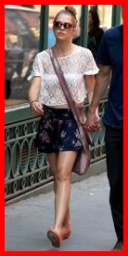}}
\end{minipage}
\\
\vspace{0.05cm}
\begin{minipage}{0.140\linewidth}
\centerline{\footnotesize{Proposed}}
\end{minipage}
\begin{minipage}{0.140\linewidth}
\centerline{}
\end{minipage}
\begin{minipage}{0.125\linewidth}
\centerline{\includegraphics[scale=0.23]{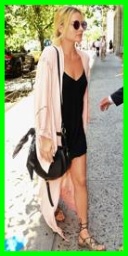}}
\end{minipage}
\begin{minipage}{0.125\linewidth}
\centerline{\includegraphics[scale=0.23]{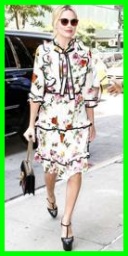}}
\end{minipage}
\begin{minipage}{0.125\linewidth}
\centerline{\includegraphics[scale=0.23]{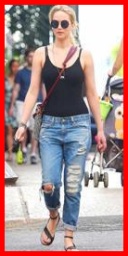}}
\end{minipage}
\begin{minipage}{0.125\linewidth}
\centerline{\includegraphics[scale=0.23]{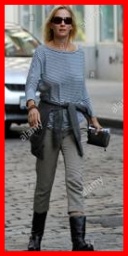}}
\end{minipage}
\begin{minipage}{0.125\linewidth}
\centerline{\includegraphics[scale=0.23]{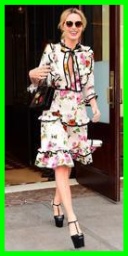}}
\end{minipage}
   \caption{Qualitative comparisons of ReIDCaps+~\citep{huang2019beyond}, SirNet~\citep{sirNet2022}, and the proposed method using the Celeb-reID dataset~\citep{huang2019beyond}. The green and red boxes denote the same and different identities compared with the person in the query images, respectively.} 
\label{fig:result_celeb}
\end{center}
\end{figure}

\begin{figure}[!t] \begin{center}
\begin{minipage}{0.140\linewidth}
\centerline{\footnotesize{Method}}
\end{minipage}
\begin{minipage}{0.140\linewidth}
\centerline{\footnotesize{Query}}
\end{minipage}
\begin{minipage}{0.125\linewidth}
\centerline{\footnotesize{Top-1}}
\end{minipage}
\begin{minipage}{0.125\linewidth}
\centerline{\footnotesize{Top-2}}
\end{minipage}
\begin{minipage}{0.125\linewidth}
\centerline{\footnotesize{Top-3}}
\end{minipage}
\begin{minipage}{0.125\linewidth}
\centerline{\footnotesize{Top-4}}
\end{minipage}
\begin{minipage}{0.125\linewidth}
\centerline{\footnotesize{Top-5}}
\end{minipage}
\\
\rule{\linewidth}{0.4pt}
\\
\vspace{0.05cm}
\begin{minipage}{0.140\linewidth}
\centerline{\footnotesize{CAL}}
\end{minipage}
\begin{minipage}{0.140\linewidth}
\centerline{\includegraphics[scale=0.23]{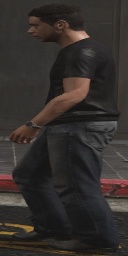}}
\end{minipage}
\begin{minipage}{0.125\linewidth}
\centerline{\includegraphics[scale=0.23]{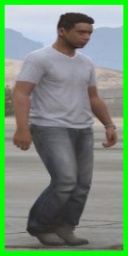}}
\end{minipage}
\begin{minipage}{0.125\linewidth}
\centerline{\includegraphics[scale=0.23]{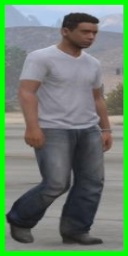}}
\end{minipage}
\begin{minipage}{0.125\linewidth}
\centerline{\includegraphics[scale=0.23]{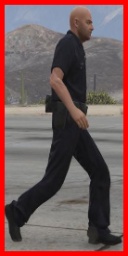}}
\end{minipage}
\begin{minipage}{0.125\linewidth}
\centerline{\includegraphics[scale=0.23]{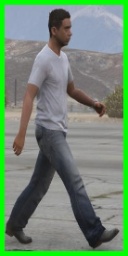}}
\end{minipage}
\begin{minipage}{0.125\linewidth}
\centerline{\includegraphics[scale=0.23]{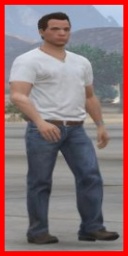}}
\end{minipage}
\\
\vspace{0.05cm}
\begin{minipage}{0.140\linewidth}
\centerline{\footnotesize{Proposed}}
\end{minipage}
\begin{minipage}{0.140\linewidth}
\centerline{}
\end{minipage}
\begin{minipage}{0.125\linewidth}
\centerline{\includegraphics[scale=0.23]{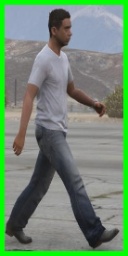}}
\end{minipage}
\begin{minipage}{0.125\linewidth}
\centerline{\includegraphics[scale=0.23]{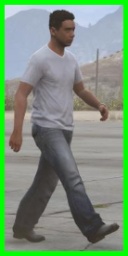}}
\end{minipage}
\begin{minipage}{0.125\linewidth}
\centerline{\includegraphics[scale=0.23]{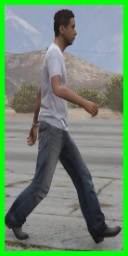}}
\end{minipage}
\begin{minipage}{0.125\linewidth}
\centerline{\includegraphics[scale=0.23]{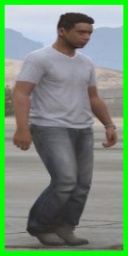}}
\end{minipage}
\begin{minipage}{0.125\linewidth}
\centerline{\includegraphics[scale=0.23]{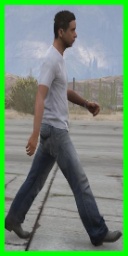}}
\end{minipage}
\\
\vspace{0.05cm}
\rule{\linewidth}{0.4pt}
\\
\vspace{0.05cm}
\begin{minipage}{0.140\linewidth}
\centerline{\footnotesize{CAL}}
\end{minipage}
\begin{minipage}{0.140\linewidth}
\centerline{\includegraphics[scale=0.23]{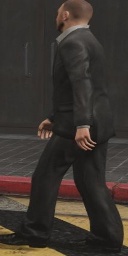}}
\end{minipage}
\begin{minipage}{0.125\linewidth}
\centerline{\includegraphics[scale=0.23]{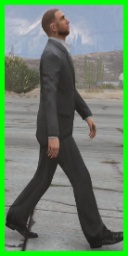}}
\end{minipage}
\begin{minipage}{0.125\linewidth}
\centerline{\includegraphics[scale=0.23]{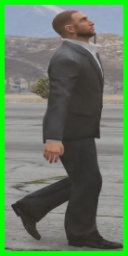}}
\end{minipage}
\begin{minipage}{0.125\linewidth}
\centerline{\includegraphics[scale=0.23]{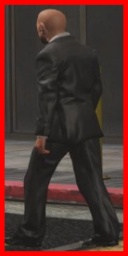}}
\end{minipage}
\begin{minipage}{0.125\linewidth}
\centerline{\includegraphics[scale=0.23]{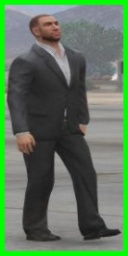}}
\end{minipage}
\begin{minipage}{0.125\linewidth}
\centerline{\includegraphics[scale=0.23]{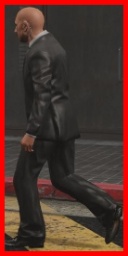}}
\end{minipage}
\\
\vspace{0.05cm}
\begin{minipage}{0.140\linewidth}
\centerline{\footnotesize{Proposed}}
\end{minipage}
\begin{minipage}{0.140\linewidth}
\centerline{}
\end{minipage}
\begin{minipage}{0.125\linewidth}
\centerline{\includegraphics[scale=0.23]{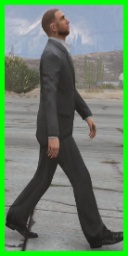}}
\end{minipage}
\begin{minipage}{0.125\linewidth}
\centerline{\includegraphics[scale=0.23]{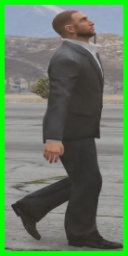}}
\end{minipage}
\begin{minipage}{0.125\linewidth}
\centerline{\includegraphics[scale=0.23]{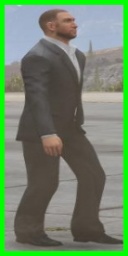}}
\end{minipage}
\begin{minipage}{0.125\linewidth}
\centerline{\includegraphics[scale=0.23]{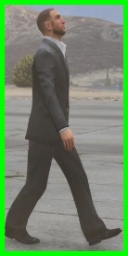}}
\end{minipage}
\begin{minipage}{0.125\linewidth}
\centerline{\includegraphics[scale=0.23]{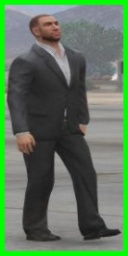}}
\end{minipage}
\\
\vspace{0.05cm}
\rule{\linewidth}{0.4pt}
\\
\vspace{0.05cm}
\begin{minipage}{0.140\linewidth}
\centerline{\footnotesize{CAL}}
\end{minipage}
\begin{minipage}{0.140\linewidth}
\centerline{\includegraphics[scale=0.23]{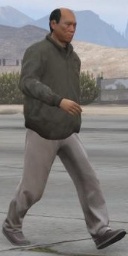}}
\end{minipage}
\begin{minipage}{0.125\linewidth}
\centerline{\includegraphics[scale=0.23]{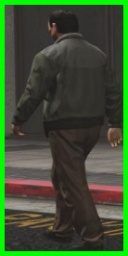}}
\end{minipage}
\begin{minipage}{0.125\linewidth}
\centerline{\includegraphics[scale=0.23]{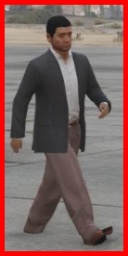}}
\end{minipage}
\begin{minipage}{0.125\linewidth}
\centerline{\includegraphics[scale=0.23]{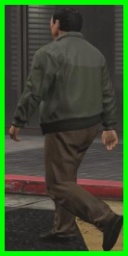}}
\end{minipage}
\begin{minipage}{0.125\linewidth}
\centerline{\includegraphics[scale=0.23]{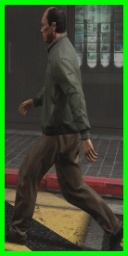}}
\end{minipage}
\begin{minipage}{0.125\linewidth}
\centerline{\includegraphics[scale=0.23]{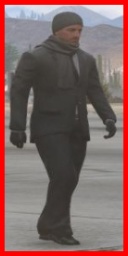}}
\end{minipage}
\\
\vspace{0.05cm}
\begin{minipage}{0.140\linewidth}
\centerline{\footnotesize{Proposed}}
\end{minipage}
\begin{minipage}{0.140\linewidth}
\centerline{}
\end{minipage}
\begin{minipage}{0.125\linewidth}
\centerline{\includegraphics[scale=0.23]{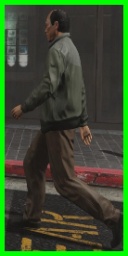}}
\end{minipage}
\begin{minipage}{0.125\linewidth}
\centerline{\includegraphics[scale=0.23]{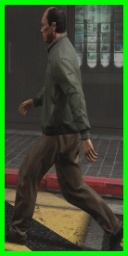}}
\end{minipage}
\begin{minipage}{0.125\linewidth}
\centerline{\includegraphics[scale=0.23]{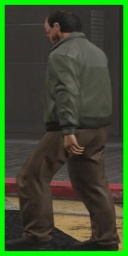}}
\end{minipage}
\begin{minipage}{0.125\linewidth}
\centerline{\includegraphics[scale=0.23]{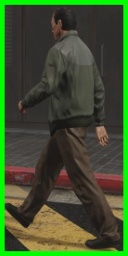}}
\end{minipage}
\begin{minipage}{0.125\linewidth}
\centerline{\includegraphics[scale=0.23]{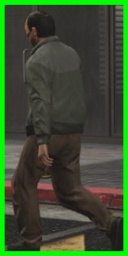}}
\end{minipage}
\\
\vspace{0.05cm}
\rule{\linewidth}{0.4pt}
\\
\vspace{0.05cm}
\begin{minipage}{0.140\linewidth}
\centerline{\footnotesize{CAL}}
\end{minipage}
\begin{minipage}{0.140\linewidth}
\centerline{\includegraphics[scale=0.23]{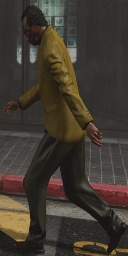}}
\end{minipage}
\begin{minipage}{0.125\linewidth}
\centerline{\includegraphics[scale=0.23]{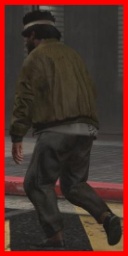}}
\end{minipage}
\begin{minipage}{0.125\linewidth}
\centerline{\includegraphics[scale=0.23]{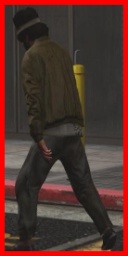}}
\end{minipage}
\begin{minipage}{0.125\linewidth}
\centerline{\includegraphics[scale=0.23]{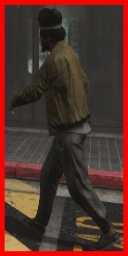}}
\end{minipage}
\begin{minipage}{0.125\linewidth}
\centerline{\includegraphics[scale=0.23]{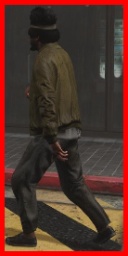}}
\end{minipage}
\begin{minipage}{0.125\linewidth}
\centerline{\includegraphics[scale=0.23]{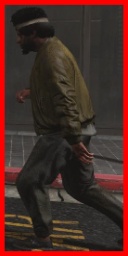}}
\end{minipage}
\\
\vspace{0.05cm}
\begin{minipage}{0.140\linewidth}
\centerline{\footnotesize{Proposed}}
\end{minipage}
\begin{minipage}{0.140\linewidth}
\centerline{}
\end{minipage}
\begin{minipage}{0.125\linewidth}
\centerline{\includegraphics[scale=0.23]{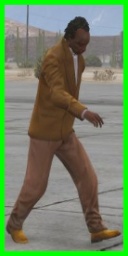}}
\end{minipage}
\begin{minipage}{0.125\linewidth}
\centerline{\includegraphics[scale=0.23]{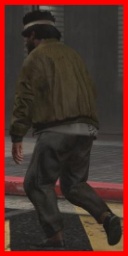}}
\end{minipage}
\begin{minipage}{0.125\linewidth}
\centerline{\includegraphics[scale=0.23]{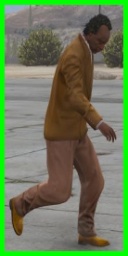}}
\end{minipage}
\begin{minipage}{0.125\linewidth}
\centerline{\includegraphics[scale=0.23]{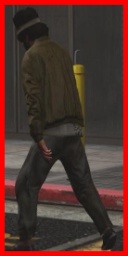}}
\end{minipage}
\begin{minipage}{0.125\linewidth}
\centerline{\includegraphics[scale=0.23]{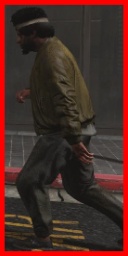}}
\end{minipage}
   \caption{Qualitative comparisons of CAL~\citep{gu2022clothes} and the proposed method using the VC-Clothes dataset~\citep{wan2020person}. The green and red boxes denote the same and different identities compared with the person in the query images, respectively.} 
\label{fig:result_vc_clothes}
\end{center}\end{figure}

\section{EXPERIMENTS AND RESULTS}
\label{sec:results}

\subsection{Dataset}
To demonstrate the effectiveness of the proposed method, we conducted experiments on three publicly available datasets: Celeb-reID~\citep{huang2019beyond}, PRCC~\citep{yang2019person}, and VC-Clothes~\citep{wan2020person}. We followed the experimental settings and dataset splits as described in previous literature~\citep{huang2019beyond, huang2019celebrities, yang2019person, wan2020person, shi2022iranet}.

\noindent \textbf{Celeb-reID}~\citep{huang2019beyond}. This dataset contains images of celebrities collected from the Internet, comprising a total of 34,186 images from 1,052 different identities. For training, 20,208 images from 632 individuals are utilized. The testing set consists of 2,972 query images and 11,006 gallery images, involving 420 identities. This dataset is suitable for evaluating the performance in scenarios involving both clothes-changing and clothes-consistent situations, as it includes images of the person with both the same and different clothes.

\noindent \textbf{PRCC}~\citep{yang2019person}. It consists of 33,698 images from 221 different identities. It contains images of individuals wearing both the same clothes (cameras A and B) and different clothes (camera C), making it suitable for evaluating both clothes-changing and clothes-consistent scenarios, similar to the Celeb-reID dataset~\citep{huang2019beyond}. It is split into 150 identities for training and 71 identities for testing. 17,896 images are utilized for training. For the clothes-changing scenario, the query dataset consists of 3,543 images from camera C, while the gallery dataset consists of 3,384 images from camera A. For the clothes-consistent scenario, the query dataset consists of 3,873 images from camera B, and the gallery dataset consists of 3,384 images from camera A. Following previous literature~\citep{shi2022iranet}, we did not utilize the 5,002 images in the validation set.

\noindent \textbf{VC-Clothes}~\citep{wan2020person}. This synthetic dataset is created using Grand Theft Auto V (GTA5) and contains 19,060 images from 512 different identities. It is split into 9,449 images from 256 identities for training and the remaining images from 256 identities for testing. The testing set consists of 1,020 query images and 8,591 gallery images.

\subsection{Experiments}
\noindent \textbf{Evaluation metrics}.
For quantitative evaluation, we employ two commonly used metrics: mean Average Precision (mAP) and Cumulative Matching Characteristics (CMC)-based top-$k$ accuracy (Rank-$k$), following previous works~\citep{huang2019beyond, huang2019celebrities, yang2019person, wan2020person}. In both cases, cosine distance $d$ is computed between the feature vector $\vf^{q}$ from a query image and the vectors $\vf^{g}$ from the gallery images, as described in~\eref{eq:cosine_distance}.

Rank-$k$ accuracy is then calculated by comparing the person in the query image to the identities of the top-$k$ ranked gallery image(s). If any of the top-$k$ ranked gallery image(s) contains the query person, the Rank-$k$ metric considers it as a correct match for the query image. The Rank-$k$ accuracy of a model is computed by measuring the proportion of query images that the model retrieves correctly.

\begin{table*}[!t]
\centering
\begin{minipage}{1\linewidth}
\caption{Quantitative comparison on the PRCC dataset~\citep{yang2019person} (\%).}
\label{tab:result_prcc}
\centering
\begin{tabular}{ >{\centering}m{0.31\textwidth}| *{3}{>{\centering}m{0.1125\textwidth}|} >{\centering\arraybackslash}m{0.1125\textwidth} } 
\hline
\multirow{2}{*}{Method} & \multicolumn{2}{c|}{Cross clothes} & \multicolumn{2}{c}{Same clothes} \\
\cline{2-5}
& Rank-1 & mAP & Rank-1 & mAP \\
\hline\hline 
HACNN~\citep{li2018harmonious} 	& 21.8 & - & 82.5 & - \\  			%% short-term + SPT
MGN~\citep{wang2018learning} 	& 33.8 & 35.9 & 99.5 & \underline{98.4} \\  	%% short-term + IRANet 
SPT~\citep{yang2019person} 		& 34.4 & - & 64.2 & - \\  			%% original paper
GI-ReID~\citep{jin2022cloth} 	& 37.6 & - & 86.0 & - \\  			%% original paper + IRANet
CASE-Net~\citep{li2021learning} 	& 39.5 & - & 71.2 & - \\  			%% original paper
HPM~\citep{fu2019horizontal} 	& 40.4 & 37.2 & 99.4 & 96.9 \\  	%% short-term + IRANet 
PCB~\citep{sun2018beyond} 		& 41.8 & 38.7 & 99.8 & 97.0 \\  	%% short-term + IRANet 
AFD-Net~\citep{xu2021adversarial} & 42.8 & - & 95.7 & - \\  			%% original paper
LightMBN~\citep{herzog2021lightweight}	& 50.0 & 50.7 & \textbf{100.0} & \textbf{99.4} \\  %% short-term + IRANet 
RCSANet~\citep{huang2021clothing} 		& 50.2 & 48.6 & \textbf{100.0} & 97.2 \\  			%% original paper
FSAM~\citep{hong2021fine} 				& 54.5 & - & 98.8 & - \\  								%% original paper
IRANet~\citep{shi2022iranet}		& \underline{54.9} & \underline{53.0} & 99.7 & 97.8 \\  						%% original paper
\hline
\textbf{Proposed} 		 & \textbf{62.2} & \textbf{60.0} & \underline{99.9} & \underline{98.4} \\
\hline
\end{tabular}
\end{minipage}
\end{table*}

\begin{table*}[!t]
\centering
\begin{minipage}{1\linewidth}
\caption{Quantitative comparison on the VC-Clothes dataset~\citep{wan2020person} (\%).}
\label{tab:result_vc_clothes}
\centering
\begin{tabular}{ >{\centering}m{0.23\textwidth}| *{2}{>{\centering}m{0.0825\textwidth}|>{\centering}m{0.0825\textwidth}|} >{\centering}m{0.0825\textwidth}| >{\centering\arraybackslash}m{0.0825\textwidth} } 
\hline
 & \multicolumn{2}{c|}{General} &\multicolumn{2}{c|}{Cross clothes} & \multicolumn{2}{c}{Same clothes} \\
Method & \multicolumn{2}{c|}{(all cams)} &\multicolumn{2}{c|}{(cam3 \& cam4)} & \multicolumn{2}{c}{(cam2 \& cam3)} \\
%  & \multicolumn{2}{c|}{} & \multicolumn{2}{c|}{} & \multicolumn{2}{c}{}\\
\cline{2-7}
& Rank-1 & mAP & Rank-1 & mAP & Rank-1 & mAP \\
\hline\hline 
MDLA~\citep{qian2017multi} 	& 88.9 & 76.8 & 59.2 & 60.8 & 94.3 & 93.9 \\  %% short-term + CAL
PCB~\citep{sun2018beyond} 	& 87.7 & 74.6 & 62.0 & 62.2 & 94.7 & 94.3 \\  %% short-term + CAL
PA~\citep{suh2018part} & 90.5 & 79.7 & 69.4 & 67.3 & 93.9 & 93.4 \\  %% short-term + CAL
FSAM~\citep{hong2021fine} 	& - & - & 78.6 & 78.9 & 94.7 & 94.8  \\  %% original paper
3DSL~\citep{chen2021learning} & - & - & 79.9 & 81.2 & - & - \\    %% original paper
CAL~\citep{gu2022clothes} 	& \underline{92.9} & \underline{87.2} & \underline{81.4} & \underline{81.7} & \underline{95.1} & \textbf{95.3}  \\  %% original paper
% LightMBN~\citep{herzog2021lightweight} & 93.0 & 87.3 & 82.2 & 83.2 & \underline{95.6} & \textbf{95.5}  \\
\hline
\textbf{Proposed} 	& \textbf{94.2} & \textbf{90.2} & \textbf{86.4} & \textbf{86.1} & \textbf{95.6} & \underline{95.2}  \\
\hline
\end{tabular}
\end{minipage}
\end{table*}

\noindent \textbf{Optimization}.
Both the OSNet~\citep{zhou2019omni} and the HRNet-W32~\citep{sun2019deep} backbones are initialized with weights pretrained on the ILSVRC dataset~\citep{krizhevsky2012imagenet}. The learning rates are scheduled using the cosine annealing method~\citep{cosineannealing2016} along with the warming-up strategy~\citep{warmingup2019}. The learning rate is initialized by $6 \times 10^{-5}$ and is linearly increased to $6 \times 10^{-4}$ during the initial 10 epochs. It is then decreased to $6 \times 10^{-7}$ over the next 150 epochs based on a cosine function. To increase the diversity of training images, horizontal flipping and random erasing~\citep{randomErasing2020} are utilized as data augmentation techniques. At each iteration, the model's parameters are updated using 42 images from 6 different identities.

\subsection{Results}
For the Celeb-reID dataset~\citep{huang2019beyond}, the quantitative results are shown in~\tref{tab:result_celeb}. The results are compared using three metrics (Rank-1, Rank-5, and mAP) following previous literature~\citep{shi2022iranet, li2021learning}. The proposed method achieves 69.0, 82.7, and 23.6 in Rank-1, Rank-5, and mAP, respectively. Compared with the previous state-of-the-art methods~\citep{shi2022iranet, li2021learning}, the proposed method achieves an absolute improvement of 2.6\%, 4.0\%, and 4.6\% in Rank-1, Rank-5, and mAP, respectively.

For the PRCC dataset~\citep{yang2019person}, Rank-1 and mAP are computed and presented in~\tref{tab:result_prcc}, following previous works~\citep{yang2019person, shi2022iranet}. To align with previous literature, we report results separately for clothes-changing and clothes-consistent scenarios, with two columns on the left representing the clothes-changing scenario and the other two on the rightmost representing the clothes-consistent scenario. In the clothes-changing scenario, the proposed method outperforms all previous methods, achieving an absolute improvement of 7.3\% in Rank-1 and 7.0\% in mAP compared with the previous state-of-the-art method~\citep{shi2022iranet}. In the clothes-consistent scenario, the proposed method achieves competitive performance, with only a 0.1\% difference in Rank-1 and a 1.0\% difference in mAP compared with the best scores. It is worth noting that most methods achieve high accuracies in the clothes-consistent setting.

\tref{tab:result_vc_clothes} presents the quantitative results on the VC-Clothes dataset~\citep{wan2020person}. This dataset allows the evaluation of methods on clothes-changing and clothes-consistent scenarios separately similar to the PRCC dataset~\citep{yang2019person}, while also having a combined evaluation setting. The experimental results demonstrate that the proposed method outperforms other methods in the clothes-changing scenario and the combined setting, while achieving competitive scores in the clothes-consistent scenario. Compared with the previous state-of-the-art method~\citep{gu2022clothes}, the proposed method achieves 5.0\% and 4.4\% higher accuracies in the clothes-changing scenario, and 1.3\% and 3.0\% improvement in the combined setting, respectively (measured in terms of Rank-1 and mAP). In the clothes-consistent setting, the proposed method achieves the highest score in Rank-1 and the second-highest performance in mAP.

Figures~\ref{fig:result_celeb} and~\ref{fig:result_vc_clothes} show the qualitative results on the Celeb-reID dataset~\citep{huang2019beyond} and the VC-Clothes dataset~\citep{wan2020person}, respectively. \fref{fig:result_celeb} shows the results of ReIDCaps+~\citep{huang2019beyond}, SirNet~\citep{sirNet2022}, and the proposed method. \fref{fig:result_vc_clothes} shows the results of CAL~\citep{gu2022clothes} and the proposed method. In both figures, the second column presents the query image, followed by the top-5 ranked gallery images displayed in the subsequent five column. The images are arranged from the closest to the fifth-closest to the query image. The green boxes indicate the same identities as the person in the query image, while the red boxes represent different identities. Alongside the quantitative results, the qualitative results also provide evidence that the proposed method outperforms the previous methods.

\begin{table*}[!t]
\centering
\begin{minipage}{1\linewidth}
\caption{Ablation study on the three streams in the proposed method using the Celeb-reID~\citep{huang2019beyond}, PRCC~\citep{yang2019person}, and VC-Clothes~\citep{wan2020person} datasets (\%).}
\label{tab:result_ablation}
\centering
\begin{tabular}{ >{\centering}m{0.06\textwidth}| >{\centering}m{0.06\textwidth}| >{\centering}m{0.06\textwidth}| *{8}{>{\centering}m{0.06\textwidth}|} >{\centering\arraybackslash}m{0.06\textwidth} } 
\hline
\multicolumn{3}{c|}{\multirow{2}{*}{Method}} & \multicolumn{3}{c|}{\multirow{2}{*}{Celeb-reID}} & \multicolumn{2}{c|}{\multirow{2}{*}{PRCC}} & \multicolumn{4}{c}{VC-Clothes} \\
\cline{9-12}
\multicolumn{3}{c|}{} & \multicolumn{3}{c|}{} & \multicolumn{2}{c|}{} & \multicolumn{2}{c|}{General} &\multicolumn{2}{c}{Cross clothes} \\ 
\hline
Global & Body part & Head & Rank-1 & Rank-5 & mAP & Rank-1 & mAP & Rank-1 & mAP & Rank-1 & mAP \\
\hline\hline
\checkmark & - & - & 54.9 & 70.2 & 12.9 & 31.3 & 34.4 & 91.1 & 81.9 & 77.1 & 77.5 \\
- & \checkmark & - & 53.9 & 70.9 & 14.0 & 32.3 & 34.1 & 90.3 & 80.8 & 75.9 & 76.2 \\
- & - & \checkmark & 59.3 & 75.7 & 17.7 & 56.3 & 45.1 & 88.5 & 77.7 & 75.0 & 68.4 \\
\checkmark & \checkmark & - & 58.0 & 72.7 & 15.6 & 39.1 & 39.0 & 92.5 & 83.5 & 79.1 & 80.2 \\
- & \checkmark & \checkmark & 65.0 & 80.2 & 21.4 & 61.4 & 51.0 & 92.7 & 85.3 & 82.0 & 78.9 \\
\checkmark & - & \checkmark & 66.6 & 80.0 & 21.7 & 54.7 & 52.1 & 93.9 & 86.4 & 84.6 & 84.3 \\
\checkmark & \checkmark & \checkmark & \textbf{69.0} & \textbf{82.7} & \textbf{23.6} & \textbf{62.2} & \textbf{60.0} & \textbf{94.2} & \textbf{90.2} & \textbf{86.4} & \textbf{86.1} \\
\hline
\end{tabular}
\end{minipage}
\end{table*}

\begin{table*}[!t]
\centering
\begin{minipage}{1\linewidth}
\caption{Analysis on the effects of adversarial erasing in the global and head stream using the Celeb-reID~\citep{huang2019beyond}, PRCC~\citep{yang2019person}, and VC-Clothes~\citep{wan2020person} datasets (\%).}
\label{tab:result_ablation2}
\centering
\begin{tabular}{ *{2}{>{\centering}m{0.075\textwidth}|} *{6}{>{\centering}m{0.065\textwidth}|} >{\centering}m{0.065\textwidth}|>{\centering}m{0.065\textwidth}|>{\centering\arraybackslash}m{0.065\textwidth}} 
\hline
\multicolumn{2}{c|}{\multirow{2}{*}{Erasing}} & \multicolumn{3}{c|}{\multirow{2}{*}{Celeb-reID}} & \multicolumn{2}{c|}{\multirow{2}{*}{PRCC}} & \multicolumn{4}{c}{VC-Clothes} \\
\cline{8-11}
\multicolumn{2}{c|}{} & \multicolumn{3}{c|}{} & \multicolumn{2}{c|}{} & \multicolumn{2}{c|}{General} &\multicolumn{2}{c}{Cross clothes} \\ 
\hline
Global & Head & Rank-1 & Rank-5 & mAP & Rank-1 & mAP & Rank-1 & mAP & Rank-1 & mAP \\
\hline\hline
- & - & 67.0 & 82.2 & 22.3 & 57.5 & 56.3 & 92.3 & 86.1 & 84.4 & 83.8 \\
- & \checkmark & 67.2 & 82.3 & 22.4 & 58.4 & 58.9 & 93.2 & 87.1 & 85.8 & 84.3 \\
\checkmark & - & 67.5 & 81.9 & 22.8 & 59.7 & 57.3 & 94.1 & 88.7 & 85.0 & 84.6 \\
\checkmark & \checkmark & \textbf{69.0} & \textbf{82.7} & \textbf{23.6} & \textbf{62.2} & \textbf{60.0} & \textbf{94.2} & \textbf{90.2} & \textbf{86.4} & \textbf{86.1} \\
\hline
\end{tabular}
\end{minipage}
\end{table*}

\subsection{Analysis}
\label{sec:result_analysis}
\tref{tab:result_ablation} presents the effects of utilizing or excluding the three streams in the proposed framework by evaluating them on the Celeb-reID~\citep{huang2019beyond}, PRCC~\citep{yang2019person}, and VC-Clothes~\citep{wan2020person} datasets. For the PRCC and VC-Clothes datasets, we provide experimental results for clothes-changing and general scenarios because performances in the same clothes scenarios are saturated and similar among recent state-of-the-art methods. The top three rows present the results obtained by utilizing an individual branch, while the subsequent three rows display the results obtained by employing two streams. The bottom row shows the results achieved by utilizing all three streams.

First of all, the results demonstrate that the three streams are complementary. The highest score is achieved when all three streams are employed together. Additionally, the results indicate that the head stream plays a significant role in the Celeb-reID and PRCC datasets, as face and hair information tends to be more robust compared with body or clothes information. Among the top three rows, the head stream consistently outperforms the other individual branches. However, for the VC-Clothes dataset, the individual head stream performs slightly lower than the other two streams because the dataset contains many samples without face information and with the backs of heads, as shown in the bottom row of~\fref{fig:result_vc_clothes}. Moreover, among the subsequent three rows, combining the head stream with another stream leads to better performance compared with combining the other two streams in all the datasets.

\begin{table}[!t]
\centering
\begin{minipage}{1\linewidth}
\caption{Analysis of weighting coefficients in the total loss using the Celeb-reID dataset~\citep{huang2019beyond} and the PRCC dataset~\citep{yang2019person} (\%).}
\label{tab:result_analysis_loss}
\centering
\begin{tabular}{ *{2}{>{\centering}m{0.08\textwidth}|} *{2}{>{\centering}m{0.12\textwidth}|} >{\centering}m{0.07\textwidth}| >{\centering}m{0.12\textwidth}| >{\centering\arraybackslash}m{0.07\textwidth} } 
\hline
\multicolumn{2}{c|}{Parameter} & \multicolumn{3}{c|}{Celeb-reID} & \multicolumn{2}{c}{PRCC}  \\
\hline
$\lambda_{pair}$ & $\lambda_{psd}$ & Rank-1 & Rank-5 & mAP & Rank-1 & mAP  \\
\hline\hline
0.9 & 0.09 & 68.0 & \textbf{82.7} & 23.7 & 61.7 & 57.9 \\
0.9  & 0.1 & 68.6 & 82.1 & \textbf{24.0} & 58.1 & 56.4  \\
0.9 & 0.11 & 68.2 & 81.7 & 23.4 & 59.5 & 56.7  \\
1   & 0.09 & 68.6 & 82.2 & 23.9 & 59.1 & 58.5 \\
1 & 0.1 & \textbf{69.0} & \textbf{82.7} & 23.6 & \textbf{62.2} & \textbf{60.0}  \\
1   & 0.11 & 68.6 & 82.7 & 23.7 & 60.0 & 57.5  \\
1.1 & 0.09 & 68.3 & 81.9 & 23.5 & 58.3 & 56.5 \\
1.1  & 0.1 & 68.9 & 82.1 & 23.7 & 59.5 & 55.2 \\
1.1 & 0.11 & 68.5 & 82.0 & 23.7 & 57.2 & 53.9  \\
\hline
\end{tabular}
\end{minipage}
\end{table}

\tref{tab:result_ablation2} demonstrates the advantages of applying adversarial erasing in the global and head streams, evaluated on the Celeb-reID~\citep{huang2019beyond}, PRCC~\citep{yang2019person}, and VC-Clothes~\citep{wan2020person} datasets. For the PRCC and VC-Clothes datasets, we present experimental results for scenarios involving clothes changes because performances in scenarios with the same clothes are saturated and similar across recent state-of-the-art methods. The results show that employing adversarial erasing in both streams provides the best performance, while applying it to only one of the streams still leads to improved accuracy.

\begin{table}[!t]
\centering
\begin{minipage}{1\linewidth}
\caption{Comparison in computational complexity.}
\label{tab:result_analysis_time}
\centering
\begin{tabular}{ >{\centering}m{0.54\textwidth}| >{\centering}m{0.13\textwidth}| >{\centering\arraybackslash}m{0.15\textwidth} } 
\hline
Method & FPS & \# of params. \\
\hline\hline
PCB~\citep{sun2018beyond} 			& 635 & 24.6M  \\  %% short-term + ReIDCaps
HACNN~\citep{li2018harmonious} 		& 852 & 3.5M  \\  %% short-term + ReIDCaps
MGN~\citep{wang2018learning} 		& 126 & 66.6M  \\  %% short-term + ReIDCaps
ReIDCaps~\citep{huang2019beyond} 	& 258 & 45.4M  \\  %% original paper
IS-GAN~\citep{eom2021disentangled} 	& 82 & 152.1M \\  %% original paper
SirNet~\citep{sirNet2022} 					& 88 & 263.8M  \\  %% original paper
LightMBN~\citep{herzog2021lightweight} 		& 242 & 9.5M  \\  %% short-term + IRANet
ReIDCaps+~\citep{huang2019beyond} 			& 55 & 45.4M  \\  %% original paper
\hline
Proposed excl. head detection 		& 107 & 32.6M  \\
Proposed 					 		& 23 & 96.6M  \\
\hline
\end{tabular}
\end{minipage}
\end{table}

We present an analysis of hyperparameters in the total loss in~\tref{tab:result_analysis_loss}. As mentioned in~\sref{sec:method_loss}, we set $\lambda_{pair}$ and $\lambda_{psd}$ to one and $0.1$, respectively, across all datasets. \tref{tab:result_analysis_loss} shows that this hyperparameter configuration achieves the highest Rank-1 accuracies, which is considered the most crucial metric, on both the Celeb-reID dataset~\citep{huang2019beyond} and the PRCC dataset~\citep{yang2019person}. Furthermore, the table demonstrates the robustness of the proposed method to variations in these weighting coefficients. Even with suboptimal hyperparameters, the proposed method consistently outperforms previous state-of-the-art methods.

\tref{tab:result_analysis_time} compares the inference speed and model complexity of the proposed method with those of other methods, using the Celeb-reID dataset~\citep{huang2019beyond}. All results are measured on a computer with an Nvidia GeForce RTX 3090 GPU and an Intel Core i9-10940X CPU. The batch size is set to 32 for all experiments. We present frames per second (FPS) for inference speed, calculated by dividing the total number of images in the query and gallery sets by the measured inference time. The inference time includes both feature extraction and distance calculation. Model complexity is presented as the number of parameters for each model.

We provide the computational complexity of the entire proposed method and the proposed method excluding head detection. The latter case assumes that cropped images of the head region are available in advance. The proposed method excluding head detection runs at 107 FPS with 32.6 million parameters, while the entire proposed method operates at 23 FPS with 96.6 million parameters. Accordingly, enhancing the overall computational complexity could be achieved by replacing the head detection method with a more efficient approach, making it a potential area for future work.

\section{CONCLUSION}
We have presented an effective framework for long-term person re-identification that takes into account both clothes-changing and clothes-consistent scenarios. The main contribution of this work is proposing a framework that effectively utilizes global and local information. The proposed framework comprises three streams: global, local body part, and head streams. The global stream captures identity-relevant information from the entire image, while the head stream focuses on an explicitly cropped head image. Each stream explicitly encodes the most distinct, less distinct, and average features. Additionally, the local body part stream utilizes pseudo-labels obtained through cascaded clustering to train the body part segmentation head, and it encodes features for each body part. The proposed framework is trained by backpropagating identity classification loss, pair-based loss, and pseudo body part segmentation loss. We conducted experiments using three publicly available datasets, and the results demonstrate that our proposed method outperforms the previous state-of-the-art method. 

In future, it would be valuable to investigate a continuously improving person re-identification framework. In real-world scenarios, while a robot navigates or provides a service, it continuously collects new data, including both successful and failed cases. These cases can be naturally annotated through human-robot interactions. By leveraging this additional data, the robot can continuously learn and improve its re-identification performance.

\section*{Acknowledgments}
This work was supported by the National Research Foundation of Korea (NRF) grant funded by the Korea government(MSIT) (No. 2020R1G1A1006143).

%\vspace{0.5cm}

\bibliographystyle{elsarticle-harv}
\bibliography{mybibfile}

\end{document}